\documentclass{article}

\usepackage{PRIMEarxiv}

\usepackage[utf8]{inputenc} 
\usepackage[T1]{fontenc}    
\usepackage{hyperref}       
\usepackage{url}            
\usepackage{booktabs}       
\usepackage{amsfonts}       
\usepackage{nicefrac}       
\usepackage{microtype}      
\usepackage{lipsum}
\usepackage{fancyhdr}       
\usepackage{graphicx}       
\graphicspath{{media/}}     
\usepackage{graphicx} 
\usepackage{subfigure}
\usepackage{amsmath}
\usepackage[table,xcdraw]{xcolor}
\usepackage{multirow}
\usepackage{threeparttable}
\usepackage{mathrsfs}
\usepackage{caption}
\usepackage{color,soul}
\graphicspath{{./Figure/}}

\title{
DEGAN: Time Series Anomaly Detection using Generative Adversarial Network Discriminators and Density Estimation
}

\author{
  Yueyan Gu\\
  Ph.D. Student \\
  Department of Civil and Environmental Engineering\\
  Virginia Tech \\
  Blacksburg, VA, USA\\
  yueyangu@vt.edu

  \And
  Farrokh Jazizadeh \\
  Associate Professor \\
  Department of Civil and Environmental Engineering\\
  Virginia Tech \\
  Blacksburg, VA, USA\\
  jazizade@vt.edu \\
  Corresponding Author \\
}

\begin{document}
\maketitle

\begin{abstract}
Developing efficient anomaly detection techniques is important to maintain service quality and provide early alarms in industrial systems. Many related efforts have been made based on time series data, a ubiquitous form for measuring, recording and analysing the dynamics of different processes. Unsupervised methods have been more popular in anomaly detection due to challenges associated with labeled datasets. Generative neural network methods are one class of the unsupervised approaches that are achieving increasing attention in recent years. In this paper, we have proposed an unsupervised Generative Adversarial Network (GAN)-based anomaly detection framework,  DEGAN. It relies solely on normal time series data as input to train a well-configured discriminator ($\mathscr{D}$) into a standalone anomaly predictor. In this framework, time series data is processed by the sliding window method. Expected normal patterns in data are leveraged to develop a generator ($\mathscr{G}$) capable of generating normal data patterns. Normal data is also utilized in hyperparameter tuning and $\mathscr{D}$ model selection steps. Validated $\mathscr{D}$ models are then extracted and applied to evaluate unseen (testing) time series and identify patterns that have anomalous characteristics. Kernel density estimation (KDE) is applied to data points that are likely to be anomalous to generate probability density functions on the testing time series. The segments of the testing time series with the highest relative probabilities are detected as anomalies. To evaluate the performance of the framework, we used a case study dataset of univariate acceleration time series for five miles of a Class I railroad track. We implemented the proposed approach to detect the real anomalous observations identified by operators. The results show that leveraging the framework with a convolutional neural network $\mathscr{D}$ architecture results in average best recall and precision of 80\% and 86\%, respectively, which demonstrates that a well-trained standalone $\mathscr{D}$ model has the potential to be a reliable anomaly detector. Moreover, the influence of GAN hyperparameters, GAN architectures, sliding window sizes, clustering of time series, and model validation with labeled/unlabeled data were also investigated to provide insight into their impact.
\end{abstract}


\section{Introduction}
Anomaly detection, namely the determination of data patterns deviating from the normal state of a system, is a classic problem in machine learning, where the quantity of abnormal and normal patterns is usually highly imbalanced. Popular application scenarios include but are not limited to financial systems \cite{golmohammadi2015time},  medical diagnosis \cite{quellec2016multiple}, building/infrastructure systems \cite{yan2018semi} and network security \cite{ahmed2016survey}. Motion and dynamics of industrial devices, weather and climate, personal and social economic activities, are typically represented by time series, a succession of data points over time. Therefore, time series anomaly detection is of significance in managing the overall service quality of a variety of systems and the Internet of Things \cite{cook_iot}, such as building/infrastructure systems including utility consumption (water, electricity, gas), structural health monitoring (deflection, displacement), indoor environment (temperature, air quality), etc. Although we have known the significance of anomaly detection for time series data, it remains a challenge due to its complicated temporal dependence and stochastic nature \cite{Su_Stochastic_Recurrent_Neural_Network_2019}. 

Existing time series anomaly detection methods can generally be divided into the following categories: statistical approaches (e.g., Autoregressive Model, AutoRegressive Integrated Moving Average (ARIMA) Model, Simple Exponential Smoothing), classical machine learning approaches(e.g., k-means clustering, isolation forest, one-class support vector machines, extreme gradient boosting), neural network approaches (e.g., Multiple Layer Perceptron, Convolutional Neural Networks, Long Short Term Memory network), and generative methods (e.g., Autoencoders, Generative Adversarial Networks) \cite{braei2020anomaly}. 
In addition, depending on how labeled data is used, anomaly detection methods can also be categorized into supervised and unsupervised approaches. Supervised approaches need labels for the input time series to differentiate anomalous and normal observations,
while unsupervised anomaly detection methods depend solely on unlabeled data. Owing to the inefficiency of label, time series anomaly detection is more common to be dealt with as an machine learning problem in an unsupervised paradigm. \cite{bashar2020tanogan}.
With the increasing computing power, more advanced machine learning approaches have emerged. Generative Adversarial Networks (GAN), since introduced in 2014 \cite{goodfellow2014generative}, have gained much popularity in image generation, data augmentation, and image-to-image translation areas. In recent years, it has also been applied in anomaly detection, mainly by relying on loss scores as metrics for anomalous pattern recognition. However, only limited research has been conducted on studying the potential of standalone discriminator ($\mathscr{D}$) models in GANs for anomaly detection. To this end, we have proposed an unsupervised method of GAN-based density estimation for time series, DEGAN, by learning the characteristics of the normal data patterns through training and validating GAN models on normal data observations. The $\mathscr{D}$ model is then extracted to identify patterns that have anomalous features. Kernel density estimation (KDE) is applied to generate probability density functions on the testing time series. 

The highlights of DEGAN framework are as follows: (1) it relies on a well-trained $\mathscr{D}$ model as a standalone anomaly detection model; (2) it doesn't need labeled data for training and the optimal $\mathscr{D}$ model selection; and (3) it can reach a relatively high recall and meanwhile well balance recall and precision. We have evaluated DEGAN using a real-world case study, i.e., detecting anomalous observations on a Class I  railroad track inspection dataset.

The rest of this paper is organized as follows. Section 2 introduces related research work on time series anomaly detection methods and Generative Adversarial Networks. In Section 3, we presented the DEGAN framework and elaborated on different framework design considerations. In Section 4, the case study has been introduced and the performance of the framework has been evaluated. In doing so, we have discussed the adopted performance metrics, as well as influencing factors that affect the overall performance. Finally, in Section 5, the main contributions and results of this paper are concluded.

\section{Related work}

Anomaly detection has been a popular research direction because of its value in monitoring conditions of different systems and providing timely alarms. How to choose a specific anomaly detection method usually depends on the type of data. 
Given our focus on time series anomaly detection and GAN-based frameworks, the scope of the review has been narrowed down to cover time-series-based and GAN-based anomaly detection techniques.
 
Time series anomaly detection is usually carried out in an unsupervised paradigm and could be challenging because of its noise and temporal dependencies \cite{Su_Stochastic_Recurrent_Neural_Network_2019}. As early as 1977,  a statistical approach was proposed by Tukey \cite{tukey1977exploratory} to detect anomalies on time series. Meanwhile, as noticeable progress has been achieved in developing machine learning approaches in the past few decades, many of them have been applied to anomaly detection problems. For example, k-means clustering \cite{zolhavarieh2014review} is an algorithm that can be executed on the sub-sequences of the time series dataset, which converges to $k$ centroids. The distance from a new testing sequence to its nearest centroid could be evaluated to identify the error. An anomaly can be reported when the corresponding error is higher than a preset threshold. Along the same line of distance-based techniques, in 2003, Ma et al. \cite{ma2003time} utilized One-Class Support Vector Machines (OC-SVM) to detect novelties (anomalies) in time series as outliers of the normal distribution, where the vectors were converted into a projected space. In 2012, Liu et al. \cite{liu2012isolation} proposed Isolation Forest (iForest), which isolates anomalies using binary trees, without conforming to the normal distribution. 

With increasing computing power, deep learning approaches are catching more attention in the past decade. These methods generally detect anomalies by comparing the new object with the normal distributions predicted based on given history data. Long Short Term Memory (LSTM) networks have been known as a useful tool for learning the longer-term pattern contained in sequences. In 2015, Malhotra et al. \cite{malhotra2015long} demonstrated stacked LSTM networks' use for multiple time series anomaly detection scenarios such as ECG, power demand and multi-sensor engine, where a network is trained on normal data and employed in the prediction for future steps. In 2018, Munir et al. \cite{munir2018deepant} proposed DeepAnt, a Convolutional Neural Networks (CNN) architecture, to predict expected patterns in a short future horizon. Using this CNN model, non-conforming patterns in the data are detected as anomalies. 

Although anomaly detection methods usually cater to specific problems, GAN, as an efficient tool applied to image anomaly detection \cite{deecke2018image}, has provided great inspiration for anomaly detection in time series. Sun et al. \cite{sun_time_2019} have characterized it to be similar to the decision-making process of human beings: $\mathscr{G}$ is responsible for learning from previous data and $\mathscr{D}$ functions as a relatively independent anomaly detector referring to $\mathscr{G}$’s knowledge of previous experience. In existing research and studies, GAN has facilitated anomaly detection of non-graphic data in the following two aspects:

\begin{itemize}
\item \textbf{Supervised approaches with oversampling of abnormal data:} GAN is widely used to generate convincing synthetic data in various scenarios, which could be utilized in alleviating the challenge of imbalanced data via creating synthetic anomalies \cite{douzas_effective_2018}. Intuitively, we can convert time series to images, thus treating non-graphic data oversampling as an image generation problem. Salem et al.\cite{salem_anomaly_2018} converted integer-based intrusion data into images and utilized a Cycle-GAN to oversample the anomalies, and their study proved that, with the aid of GAN as a supplementary oversampling tool, the anomaly detection performance has been improved. This could be specifically beneficial for an inherently imbalanced dataset.
\item \textbf{Unsupervised approaches via defining an anomaly score with $\mathscr{G}$ and/or $\mathscr{D}$ loss:} $\mathscr{G}$'s training loss is a measurement of the residuals between the normal/real input data and the new samples generated (reconstructed) by $\mathscr{G}$, while $\mathscr{D}$'s training loss represents the $\mathscr{D}$'s ability to distinguish real/normal from fake/abnormal. By combining both losses, $\mathscr{G}$ and $\mathscr{D}$ could be both leveraged in constructing an anomaly score to detect a sequence with anomalous patterns. Li et al. \cite{li_mad-gan_2019} proposed this idea in 2019 and applied it in unsupervised multivariate time series anomaly detection with LSTM-RNN as the $\mathscr{G}$ and $\mathscr{D}$'s base model. Similar idea has also been implemented in \cite{jiang_novel_2019},\cite{bashar2020tanogan}, \cite{geiger2020tadgan} and \cite{niu2020lstm}. The multivariate time series could also be transformed into 2D images by calculating distance matrices, which turns the problem into image anomaly detection with similarly defined GAN-based anomaly scores \cite{choi_gan-based_2020}. Moreover, some research efforts have also been made based only on $\mathscr{G}$ loss  \cite{zhou2019beatgan} \cite{jiang2019gan} \cite{geiger2020tadgan}. Overall, previous methods for GAN-based time series anomaly detection focus either on the $\mathscr{G}$ or on the combination of both the $\mathscr{G}$ and the $\mathscr{D}$, while only few efforts \cite{sun_time_2019} \cite{geiger2020tadgan} have been made to leverage the potential of an independent $\mathscr{D}$, which relies solely on $\mathscr{D}$'s classification result.
\end{itemize}

In this paper, we focus on developing a reliable framework for time series anomaly detection utilizing a standalone trained $\mathscr{D}$ model of GAN, with the aid of using a sliding window for feature extraction, data characterization (normal data generation and time series pattern clustering) and density estimation to quantify the probability of anomalous data patterns. Motivated by GANomaly framework \cite{akcay_ganomaly_2019}, which addresses anomaly detection problems in the computer vision domain with only normal images as input, our $\mathscr{G}$ solely learns the distributions of normal patterns and enables the $\mathscr{D}$ to acquire the knowledge to distinguish fake (abnormal) from real (normal) during the training process. Then the better-trained $\mathscr{D}$ is validated and identified to be used in conjunction with kernel density estimation. The proposed framework has been evaluated on a real-world dataset collected through standard procedures of railroad inspection with available labels based on a meticulous assessment by operators in real time.

\section{Methodology}
\begin{figure*}[h]
  \centering
  \includegraphics[width=12cm]{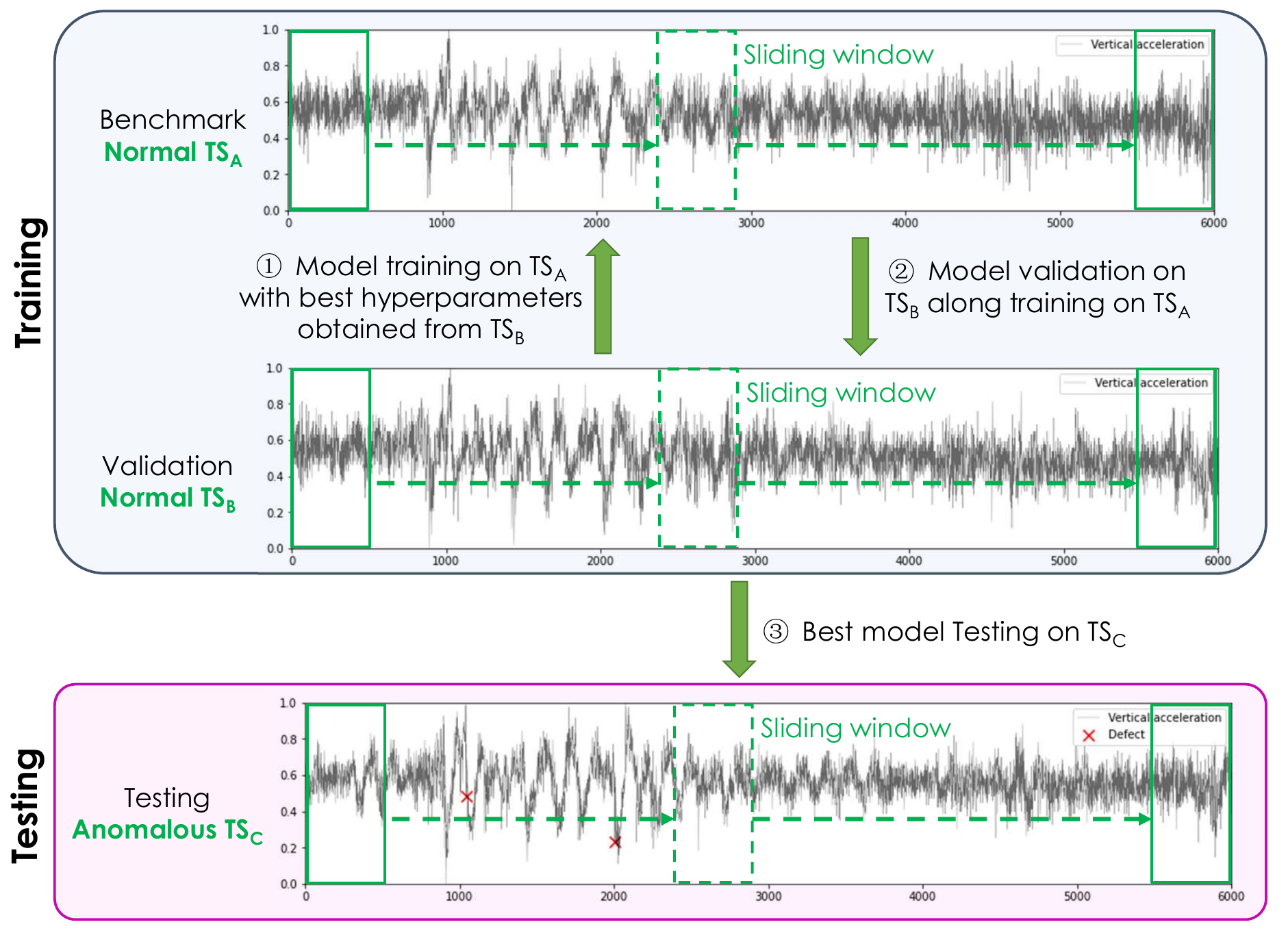}
  \caption{Time series involved in DEGAN}
  \label{fig:ts}
\end{figure*}

\begin{figure*}[h]
  \centering
  \includegraphics[width=13cm]{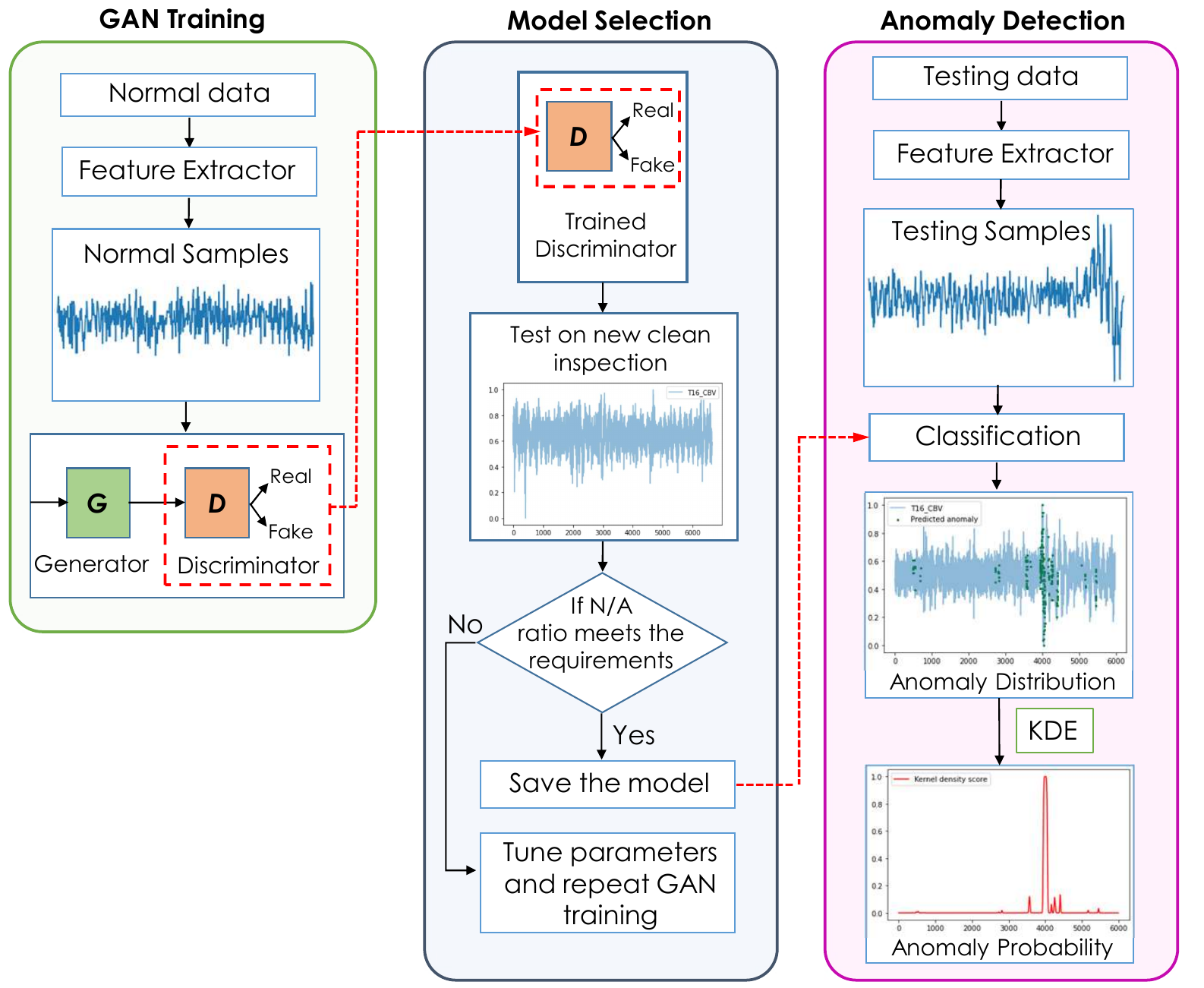}
  \caption{Framework of DEGAN}
  \label{frame}
\end{figure*}

The DEGAN framework centers around repeated time series acquired to monitor the temporal variation in the operation of a given system. These time series reflect the performance of the system and the goal is to identify when/where an anomalous event is observed. Fig.~\ref{fig:ts} shows the overall process of using repeated time series data for training, validation and testing in DEGAN, where $TS_A$ is training data, $TS_B$ is validation data and $TS_C$ is testing data. All time series are processed for subsequence extraction using a sliding window with a length of $wl$. The algorithm  relies on two segments of benchmark time series with no observed anomalies (clean) ($TS_A$ and $TS_B$) that reflect the normal behavior of the system, to train and configure the anomaly detection model. Grid search of GAN hyperparameters ($\mathscr{G}$'s and $\mathscr{D}$'s learning rates) is first carried out on $TS_B$. Then, the model is trained on $TS_A$ with those learning rates and validated on $TS_B$ along training (to decide the best epoch to stop training). The best standalone  $\mathscr{D}$ model is then used as anomaly detector on new time series (e.g., $TS_C$). 

The DEGAN framework is depicted in Fig.~\ref{frame}, which includes three main components: GAN training, $\mathscr{D}$ model selection, and probabilistic anomaly detection. These components have been described in the following subsections:

\subsection{GAN training}
 
\subsubsection{Data pre-processing}
\begin{figure}[h]
  \centering
  \includegraphics[width=8cm]{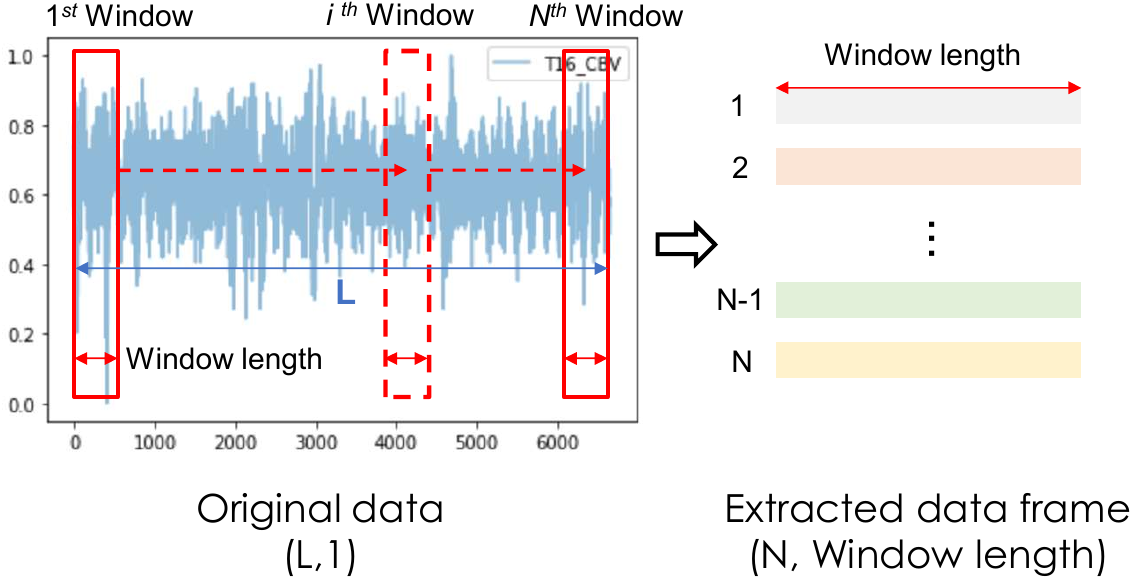}
  \caption{Subsequence extraction with the sliding window method}
  \label{sliding_window}
\end{figure}

\
As noted, all the time series are processed into subsequences (continuous segments) using a sliding window method. The trained  $\mathscr{D}$ model then uses a sliding window search to identify anomalous patterns. By applying the sliding window method, the time series is processed into a feature matrix as shown in Eq.~(\ref{eq:sliding_window}) and Fig.~\ref{sliding_window} - i.e., a time series with the shape of (\textit{L},1) is transformed into a 2D-tensor with the shape of $(N,wl)$, where $N$ is the total number of feature vectors. The sliding window length $wl$ is an important hyperparameter in this framework as it has been further discussed in Section \ref{sec: DEGAN performance using different window sizes}. Moreover, each subsequence is processed using zero-mean normalization (see Eq.~(\ref{eq:norm})).

\begin{equation}
W:=\left ( W_{1},W_{2},...,W_{N} \right)^{T}=\left ( \left(x_{1},...,x_{wl} \right),\left(x_{2},...,x_{wl+1} \right),...,\left(x_{N},...,x_{N+wl-1} \right) \right )^{T}
\label{eq:sliding_window}
\end{equation}

\begin{equation}
W:=\left ( ( W_{1}-\overline{W_{1}}),(W_{2}-\overline{W_{2}}),...,(W_{N}-\overline{W_{N}}) \right)^{T}
\label{eq:norm}
\end{equation}

\subsubsection{The GAN architecture}\
\label{sec:The GAN architecture}
A GAN network is made up of two neural networks, a generator $\mathscr{G}$ and a discriminator $\mathscr{D}$. $\mathscr{G}$ is responsible for generating synthetic (fake) data points (i.e., time series subsequences) using a random signal as input, while the $\mathscr{D}$ attempts to distinguish them from the real ones (i.e., the training set). The training process is a two-player minmax game as reflected in Eq.~(\ref{eq:GAN}). 

\begin{equation}
    \underset{G}{min}\:\underset{D}{max}~V(D,G)= \varepsilon _{x\sim p_{data}(x)}[logD(x)]+\varepsilon _{z\sim p_{z}(Z)}[log(1-D(G(z)))]
    \label{eq:GAN}
\end{equation}

In Eq.(\ref{eq:GAN}), $x$ follows the distribution of the input normal while $z$ follows a random distribution. 

\begin{figure}
  \centering
    \subfigure[$\mathscr{G}$]{              
        \includegraphics[height=4cm]{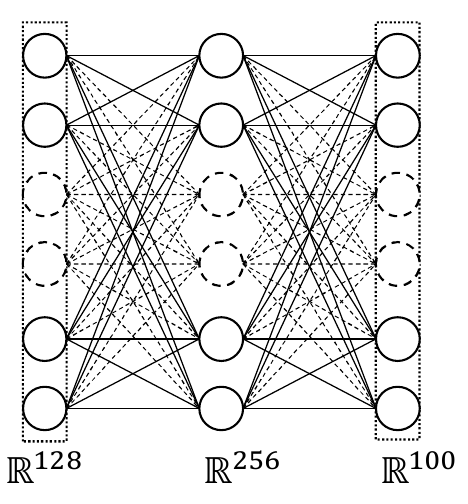}}
    \hspace{15pt}
    \subfigure[CNN-$\mathscr{D}$]{
        \includegraphics[height=4cm]{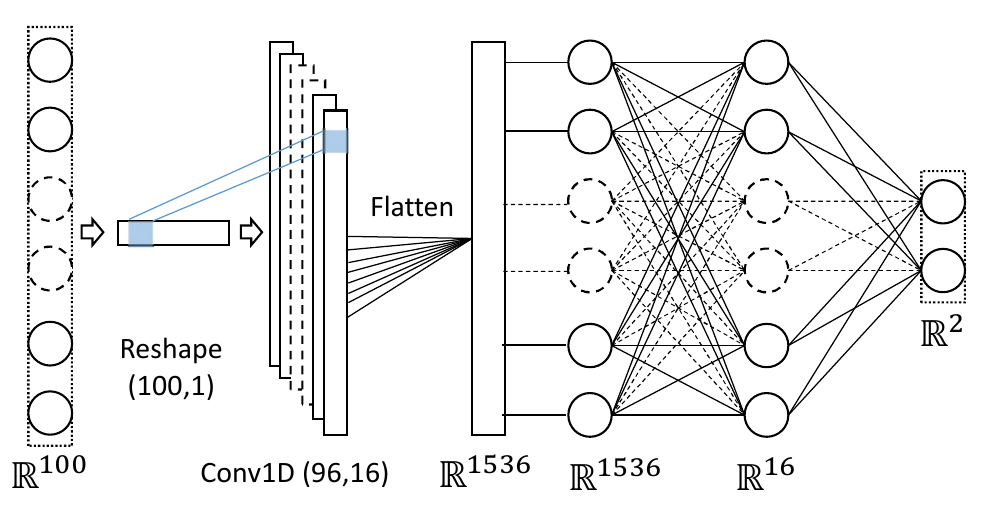}}
    \caption{Architectures of $\mathscr{G}$ and CNN-$\mathscr{D}$ ($wl$=100)}
    \label{fig:g_d_arch}
\end{figure}

The architectures of $\mathscr{G}$ and $\mathscr{D}$ employed in our framework are illustrated in Fig.~\ref{fig:g_d_arch} and summarized in Table \ref{tab:g_d_arch}. We adopted a two-layer Dense neural network as the base model of $\mathscr{G}$. The input layer is a 1d-tensor of random values drawn from a fixed standard Gaussian distribution ranging between 0 and 1 with a dimension of 128 (although it could be a different size). The two fully connected layers are followed by a Tanh activation layer. For the $\mathscr{D}$, we employed a 1d-convolutional model, CNN-$\mathscr{D}$, which consists of one convolutional layer (Conv1D) and two fully connected layers. The Conv1D layer has 16 channels of filters with a size of 5. The output of Conv1D is flattened before being fed into Dense layers. The impact of alternative GAN architectures has been discussed in Section \ref{sec: Influence of GAN architectures}.

\begin{table}[]
\centering
\caption{Summary of $\mathscr{G}$ and CNN-$\mathscr{D}$ ($wl$=100)}
\label{tab:g_d_arch}
\begin{tabular}{@{}lcllccc@{}}
                                & (a) $\mathscr{G}$          &                             &  & \multicolumn{1}{l}{} & (b) CNN-$\mathscr{D}$     & \multicolumn{1}{l}{} \\ \cmidrule(r){1-3} \cmidrule(l){5-7} 
\multicolumn{1}{c}{Layer}       & Output Shape         & \multicolumn{1}{c}{Parameter number} &  & Layer                & Output Shape  & Parameter number              \\ \cmidrule(r){1-3} \cmidrule(l){5-7} 
\multicolumn{1}{c}{InputLayer}  & (None, 128)          & \multicolumn{1}{c}{0}       &  & InputLayer           & (None, 100)   & 0                    \\
\multicolumn{1}{c}{Dense(Tanh)} & (None, 256)          & \multicolumn{1}{c}{33024}   &  & Reshape              & (None, 100,1) & 0                    \\
\multicolumn{1}{c}{Dense} & (None, 100) & \multicolumn{1}{c}{25700} &  & Conv1D (Relu) & (None, 96,16) & 96 \\
                                & \multicolumn{1}{l}{} &                             &  & Dropout              & (None, 96,16) & 0                    \\
                                & \multicolumn{1}{l}{} &                             &  & Flatten              & (None, 1536)  & 0                    \\
                                & \multicolumn{1}{l}{} &                             &  & Dense (Tanh)         & (None, 16)    & 24592                \\
                                & \multicolumn{1}{l}{} &                             &  & Dense (Sigmoid)      & (None, 2)     & 34                   \\ \cmidrule(r){1-3} \cmidrule(l){5-7} 
\end{tabular}
\end{table}

\subsubsection{Motif-based GAN training}
\begin{figure}
  \centering
    \subfigure[Dispersed sinusoidal waves]{              
        \includegraphics[width=5cm]{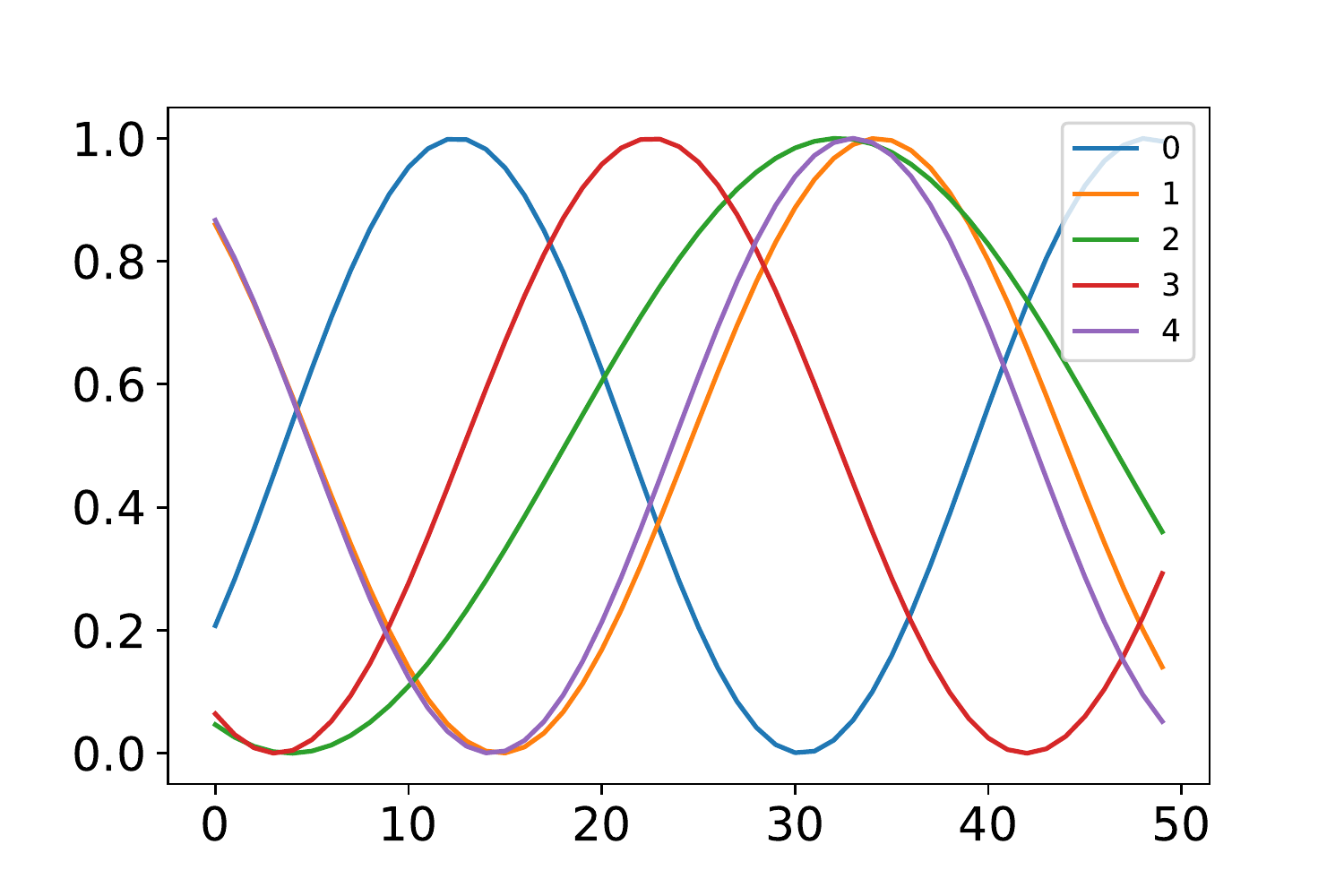}}
    \subfigure[Clustered sinusoidal waves]{
        \includegraphics[width=5cm]{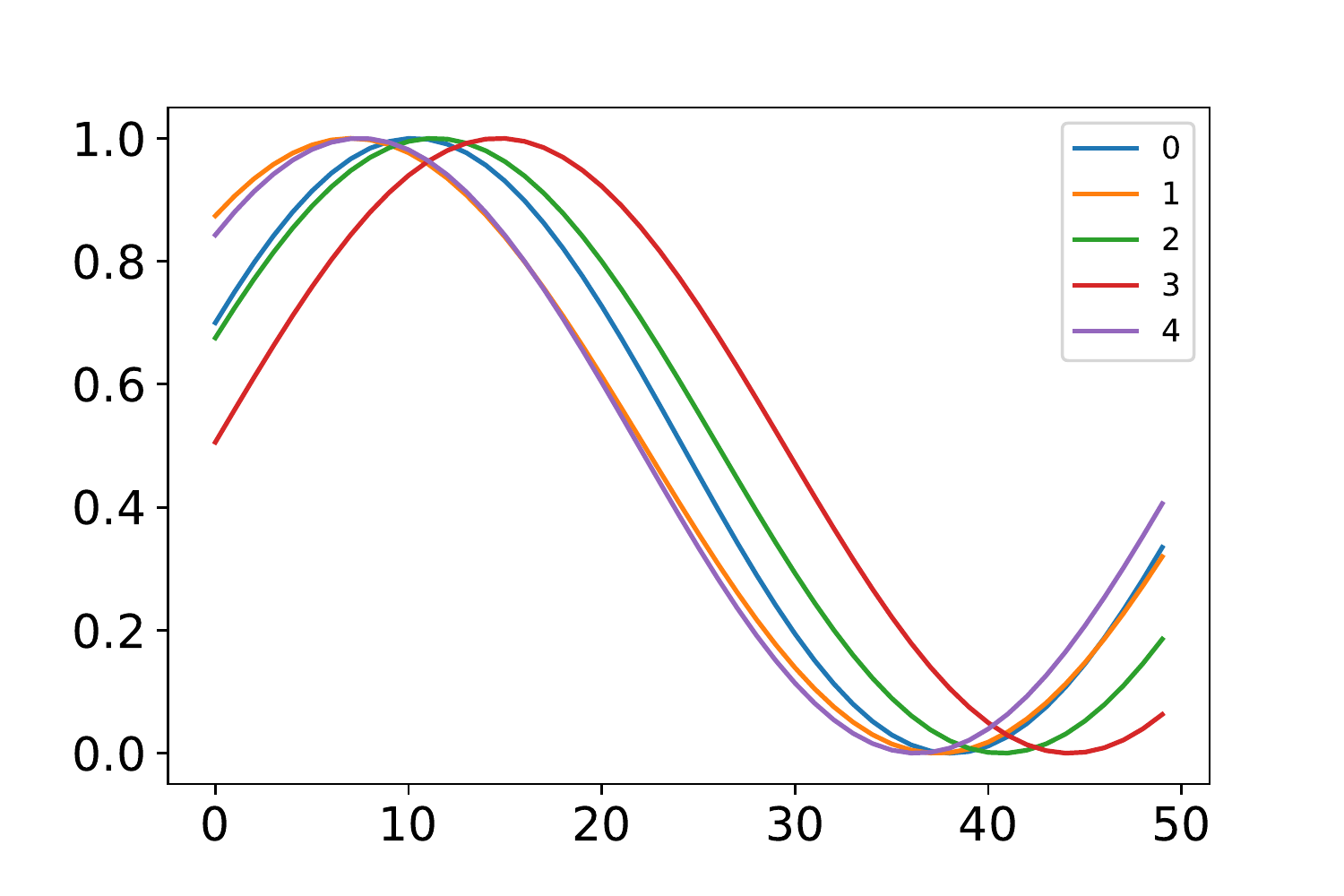}}
    \caption{Sinusoidal waves for GAN training}
    \label{sinewaves}
\end{figure}

\begin{figure}
  \centering
    \subfigure[Dispersed sinusoidal waves]{              
        \includegraphics[width=5cm]{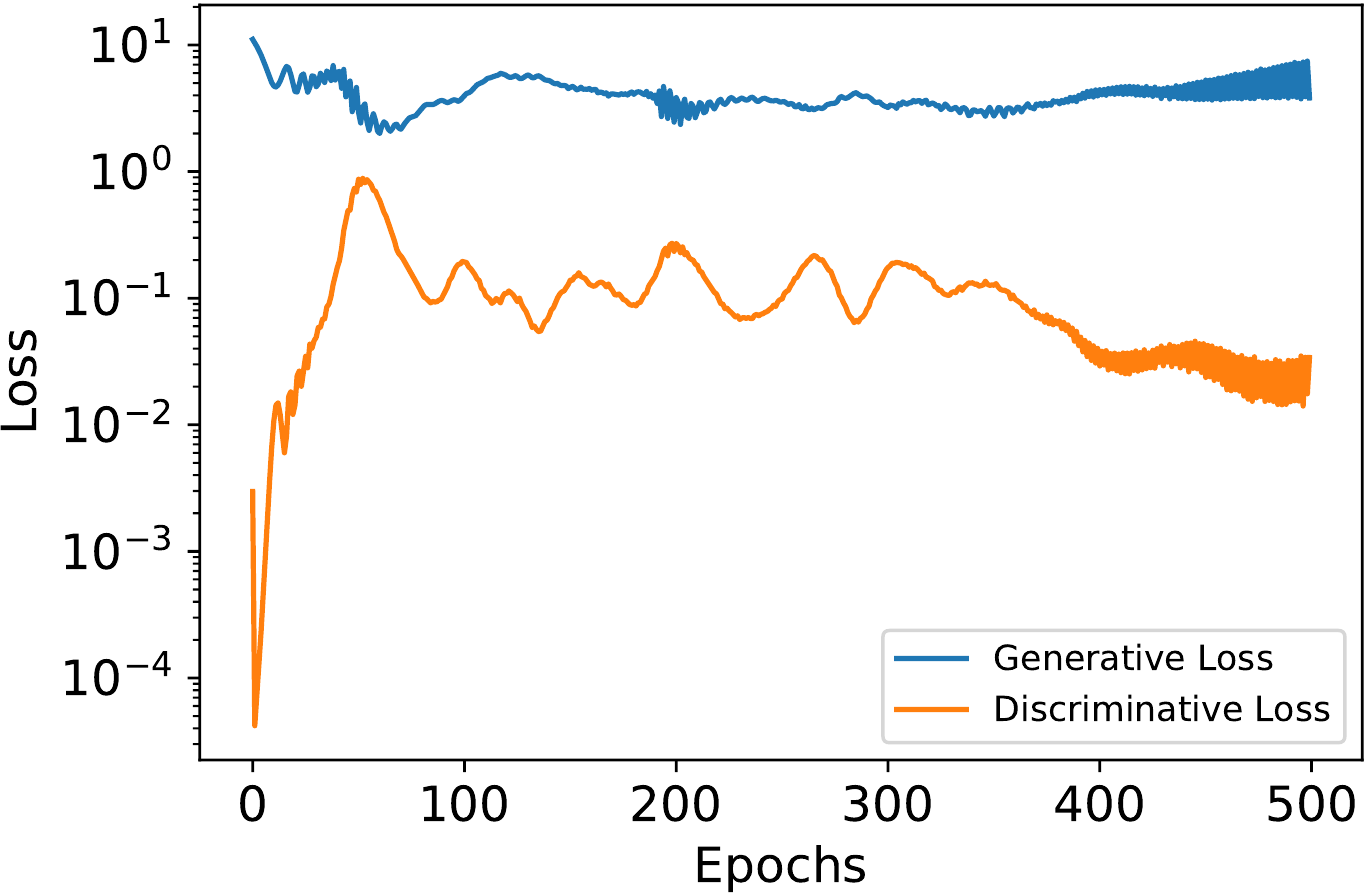}}
    \subfigure[Clustered sinusoidal waves]{
        \includegraphics[width=5cm]{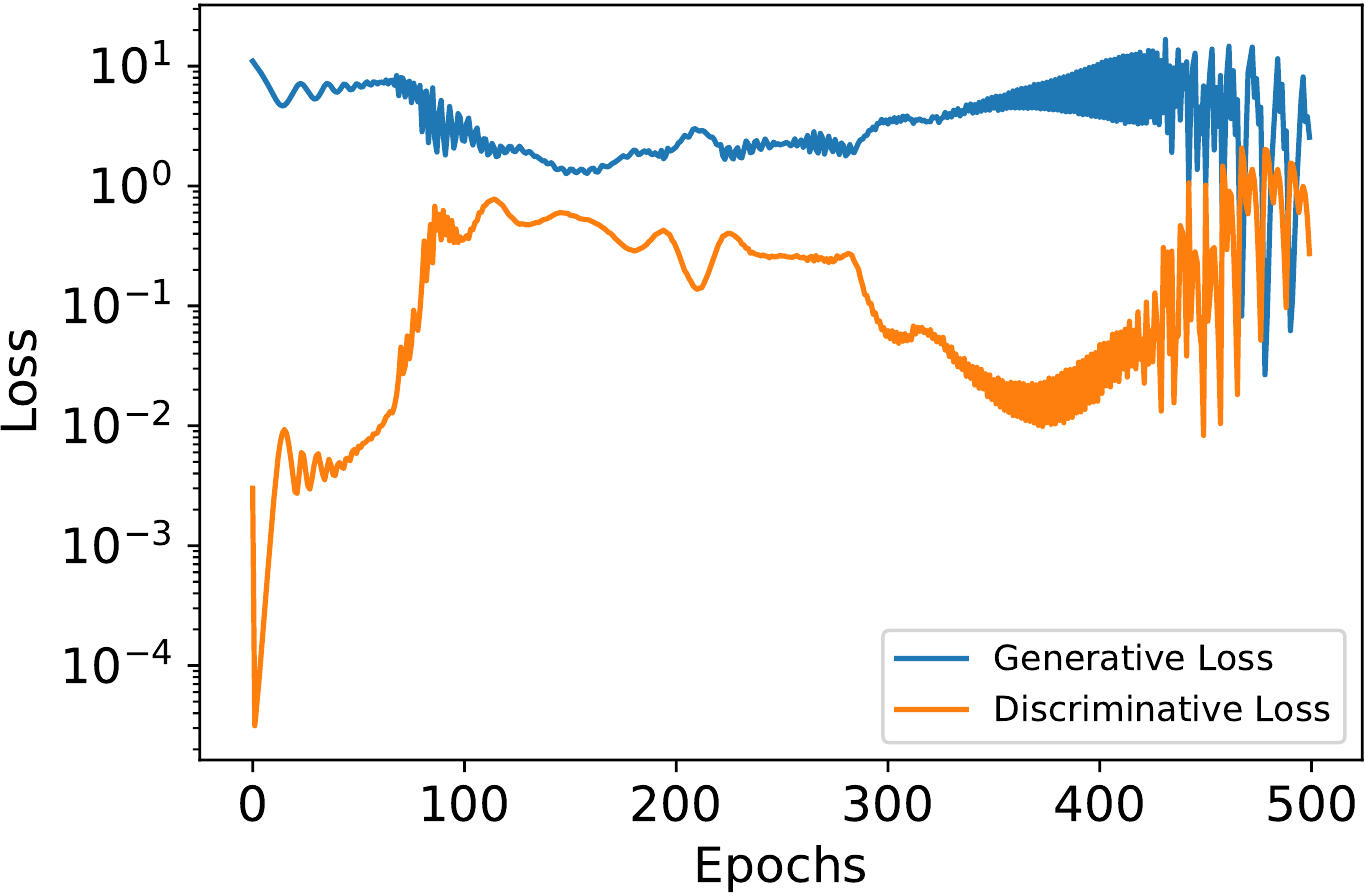}}
    \caption{Sinusoidal wave GAN training loss}
    \label{sine_training}
\end{figure}

\begin{figure*}[h]
  \centering
  \includegraphics[width = 13cm ]{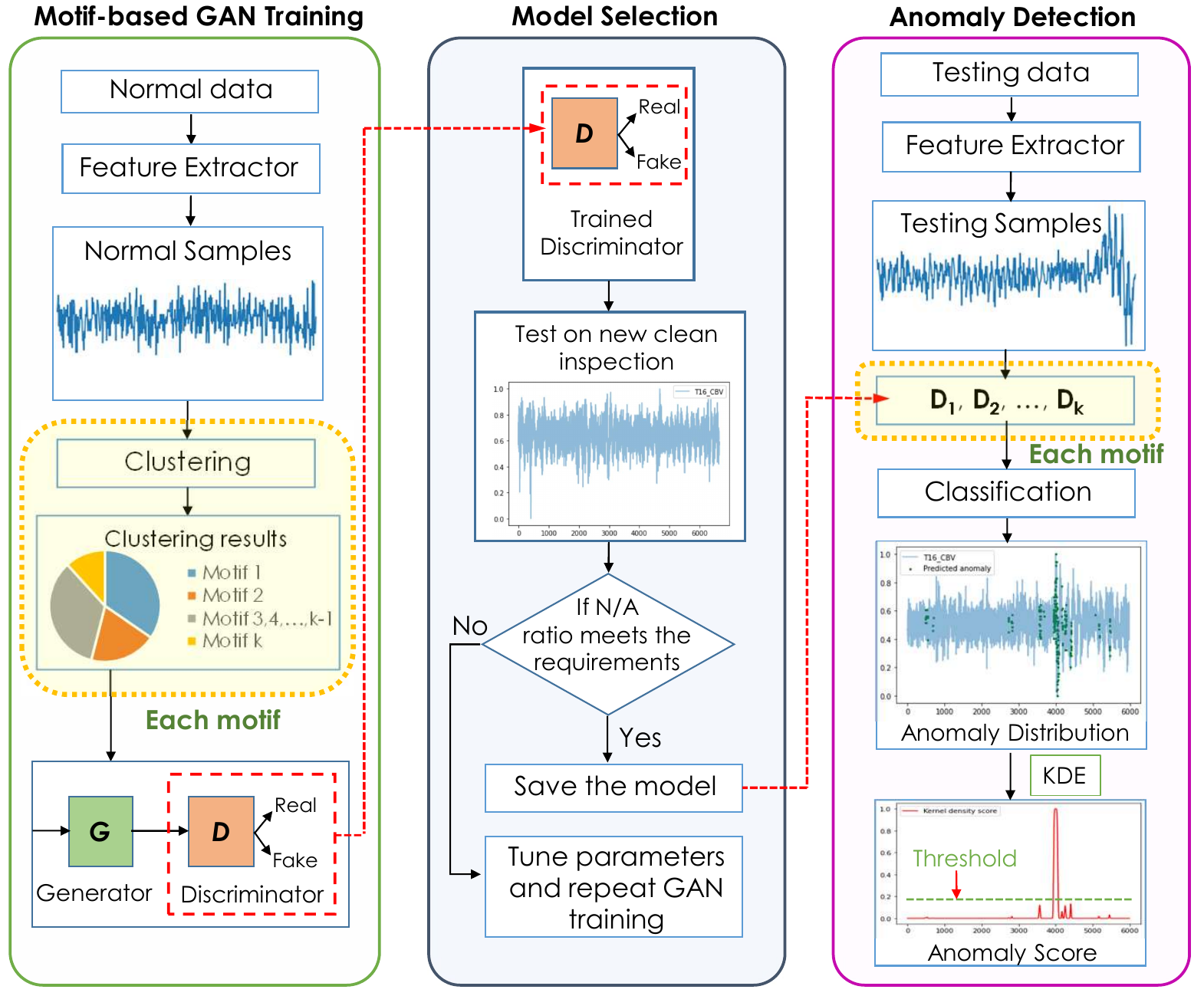}
  \caption{Framework of motif-based GAN training}
  \label{motif_GAN}
\end{figure*}

\
\label{sec: methodology - Motif-based GAN training}
In the GAN training component, we have also considered the potential of a clustering (k-means) step that enables GAN training on common distinctive normal patterns (i.e., motifs) - training different GAN models for different clusters. The addition of this step was motivated by an observational assessment of the GAN framework on a set of sinusoidal waves. To this end, a GAN model was trained on 10000 sinusoidal waves. Then, its $\mathscr{D}$ model was used to detect linear lines out of sinusoidal waves. The sinusoidal waves for both training and testing had different frequencies and phase shifts. 
As shown in Fig.~\ref{sinewaves}, two training sets of dispersed and clustered sinusoidal waves were used. As Fig.~\ref{sine_training} illustrates, for the clustered set, the $\mathscr{G}$ and the $\mathscr{D}$ tangled more closely during the training process. 
It turned out that the $\mathscr{D}$ considerably outperforms on the clustered data (Table~\ref{sine_performance}). Fig.~\ref{motif_GAN} shows the added clustering step into the framework, where the closest centroid for each testing sample is used during the anomaly detection process. The efficacy of the clustering step has been discussed in Section \ref{sec: Influence of clustering}. 

\begin{table*}[]
\centering
\begin{threeparttable}

\caption{Performance of standalone $\mathscr{D}$ on sinusoidal-waves anomaly detection}

\label{sine_performance}
\begin{tabular}{ccc}
\hline
                     & Precision & Recall \\ \hline
Dispersed sinusoidal waves & 1.0       & 0.41   \\
Clustered sinusoidal waves & 1.0       & 0.97   \\ \hline

\end{tabular}
\begin{tablenotes}
\footnotesize
\item (1) The trained $\mathscr{D}$ (at 500 epochs) was tested on a set of 500 sinusoidal waves and 50 linear lines (i.e., Imbalance Ratio = 10).
\item (2) The values in this table are the average of three runs.
\end{tablenotes}
\end{threeparttable}
\end{table*}

\subsection{$\mathscr{D}$ model selection}
\label{sec:Discriminator model selection}
An effective model validation scheme enables the early stop of machine learning model training, which helps improve the computing efficiency. 
Quantitative GAN evaluation approaches for image generation are getting more attention in recent years \cite{borji2019pros}. In our proposed framework, we also require a quantitative validation to stop the training and extract the validated $\mathscr{D}$ model. However, different from those evaluation methods in computer vision that focus on the performance of the $\mathscr{G}$, we put the emphasis on the classifying ability of the $\mathscr{D}$. Therefore, by model selection, we are specifically referring to the selection of the $\mathscr{D}$ model.

Using a labeled validation set to select a model is an intuitive and straightforward method when there is sufficient labeled data. Owing to the insufficiency of labeled data in time series anomaly detection problems, we have sought to rely solely on normal data for the model evaluation and selection instead. To this end, we proposed the Normal/Abnormal ($N/A$) ratio metric (see Eq. (\ref{NA})) by evaluating the trained $\mathscr{D}$ on a second clean time series (i.e., $TS_B$ in Fig.~\ref{fig:ts}). The intrinsic idea behind this metric is that, ideally, we should have a $\mathscr{D}$ that detects no anomaly on a clean time series. Thus, if the $N/A$ ratio is too low, the $\mathscr{D}$ has not yet been well-trained and  further training is needed. On the other hand, if the $N/A$ ratio is high, which means it correctly classifies most data points as normal, the $\mathscr{D}$ could thus be used as a qualified anomaly detector. In some cases, the $N/A$ ratio could reach infinity, when data points are all predicted as normal. In our pre-experiments, it turned out that such models are too inclusive in processing the testing time series and predict every data point as normal. As such, in the proposed framework, a validated $\mathscr{D}$ model is selected when the $N/A$ ratio reaches its highest (but not infinity) or the training reaches a maximum training epoch. In our case study, the best $\mathscr{D}$ was usually obtained between the 40th and 60th epochs and we set the maximum number of training epochs to be 100. The feasibility of this model selection method is proved in Section \ref{sec: Impact of labeled vs. unlabeled Discriminator validation} by comparing it with a supervised validation method.

\begin{equation}
N/A= \frac{\# Predicted\ Normal\ Points}{\# Predicted\ Anomalous Points}
\label{NA}
\end{equation}

\subsection{Anomaly Detection}
In the anomaly detection component (Fig.~\ref{frame}), upon normalization, each test data point is evaluated by the selected  $\mathscr{D}$ model. A $\mathscr{D}$ score threshold of > 0.5 is used to detect anomalies. As mentioned before, the anomaly detector model uses a sliding window search. Therefore, anomalous patterns produce multiple continuous suspicious data points on the test time series (see examples in Fig.~\ref{test_results}). Thus, identifying the specific point of anomaly can be challenging. To address this uncertainty, we employ kernel density estimation (KDE)\cite{parzen1962estimation}, which is used to estimate a probability density function of anomalous patterns based on the finite identified anomalous data points 
($x_1$, $x_2$, …, $x_n$).
We can estimate its density function $f$ with Eq.(\ref{KDE}):

\begin{equation}
\widehat{f}_{h}\left ( x \right )=\frac{1}{n}\sum_{i=1}^{n}K_{h}\left ( x-x_{i} \right )=\frac{1}{nh}\sum_{i=1}^{n}K\left ( \frac{x-x_{i}}{h} \right )
\label{KDE}
\end{equation}

In Eq. (\ref{KDE}), $K$ is a non-negative kernel function, and bandwidth $h > 0$ is a parameter of smoothing level. A smaller bandwidth may lead to less smooth kernel density curves and too many anomaly predictions, while a larger bandwidth may result in too few anomaly predictions thus missing the real defects. Thus, KDE bandwidth is an important parameter in the evaluation process, which should be tuned depending on the context of the anomaly detection problem - i.e., time series data characteristics, sliding window length, etc. In addition to the bandwidth, the kernel function also plays a key role. Common functions include uniform, triangular, biweight, triweight, Epanechnikov, normal, etc. Based on a few pre-experiments, considering the prediction resolution required in the case study, we decide to employ an estimator with a Gaussian kernel function and a bandwidth of 50 in our case study. 

The probability distribution function from the KDE algorithm is processed using a peak selection algorithm (function \textit{find\_peaks} from Python library \textit{scipy.signal}), which finds all local maxima by a simple comparison of neighboring values. It allows users to define a number of parameters including a minimum height for selected peaks. To automatically identify this threshold from the data itself, we have used the histogram of the probability distribution obtained from the KDE process. To this end, we have used the size of the middle bin in a 21-bin histogram as the height constraint. Alternatively, we could rely on a predefined threshold which can be one of the hyperparameters of the proposed framework.

\section{Case study: Anomaly detection of a railroad track}
To verify the efficacy of DEGAN, we carried out anomaly detection analysis for a 5-mile segment of a Class I railroad track. We evaluated the precision and recall of our predicted anomalies by comparing the predictions with the manually-labeled anomalies (taken as ground truth) in the dataset. 
In what follows, we first introduce our case study dataset and evaluation metrics calculation, and then discuss the results and the impact of GAN hyperparameters, window sizes, clustering, GAN architectures, and the model validation mode (supervised vs. unsupervised). 

\subsection{Introduction of the dataset}

\textbf{Dataset description:} The dataset contains vertical acceleration time series collected by using a track geometry car through regular maintenance inspections. The acceleration data is the average of the acceleration for each foot (30.48 cm) of the track. Through computing processes on-board, the collected raw data has been processed into time series of data samples across the track. In addition to the acceleration data, each sample on the time series data includes a timestamp and the distance passed a milepost. In this dataset, we have time series data from five inspections for five miles of a track with more than 5000 samples per mile. The total number of anomalies is only 39 (see Table \ref{tab:defect_count}).

\textbf{Time series segmentation:}  In presenting the results, we have referenced mileposts (MP) from MP1 to MP5 in the text as shown in Table \ref{tab:defect_count}. We have divided the data to cover one-mile segments as our unit of evaluation. Since the data collection and reporting for this dataset is based on mileposts, we have adopted that unit for our experiments although other units could be considered as well. Moreover, different sections of a railroad track might have different environments and conditions that call for dividing the track into multiple segments. The computational cost is another factor that should be considered for selecting the time series anomaly detection evaluation unit.

\begin{table*}[]
\caption{Number of labeled anomlaies per milepost per inspection}
\label{tab:defect_count}
\centering
\begin{threeparttable}

\begin{tabular}{ccccccc}
\hline
\multirow{2}{*}{Inspection} & \multicolumn{5}{c}{Milepost (MP)} & \multirow{2}{*}{Sum per inspection} \\
                            & 1  & 2  & 3   & 4  & 5   &                                     \\ \hline
1                           & 0  & 0  & 5   & 3  & 9   & 17                                  \\
2                           & 1  & 1  & 6   & 4  & 4   & 16                                  \\
3$^*$                           & 0  & 0  & 0   & 0  & 0   & 0                                   \\
4                           & 3  & 2  & 1   & 0  & 0   & 6                                   \\
5$^*$                           & 0  & 0  & 0   & 0  & 0   & 0                                   \\ \hline
Sum per mile                & 4  & 3  & 12  & 7  & 13  & Total = 39                          \\ \hline
\end{tabular}
\begin{tablenotes}
\footnotesize
\item $^*$ These are clean inspections containing no anomalies.
\end{tablenotes}
\end{threeparttable}
\end{table*}

\textbf{Repeated inspections:} The dataset includes five inspection runs (see Table \ref{tab:defect_count}), among which at least two are clean inspections, so that one could be used for GAN training and another one could be utilized for hyperparameter tuning and $\mathscr{D}$ model selection. 

\textbf{Labeled anomalies:} This dataset includes human-provided anomaly information, which can be viewed as labels that we leveraged in testing and evaluation. As the proposed framework is in an unsupervised paradigm, labeled data is only used in the testing phase to evaluate the framework's performance. 

\begin{figure*}
  \centering
    \subfigure[the $1^{st}$ window]{              
        \includegraphics[width=4cm]{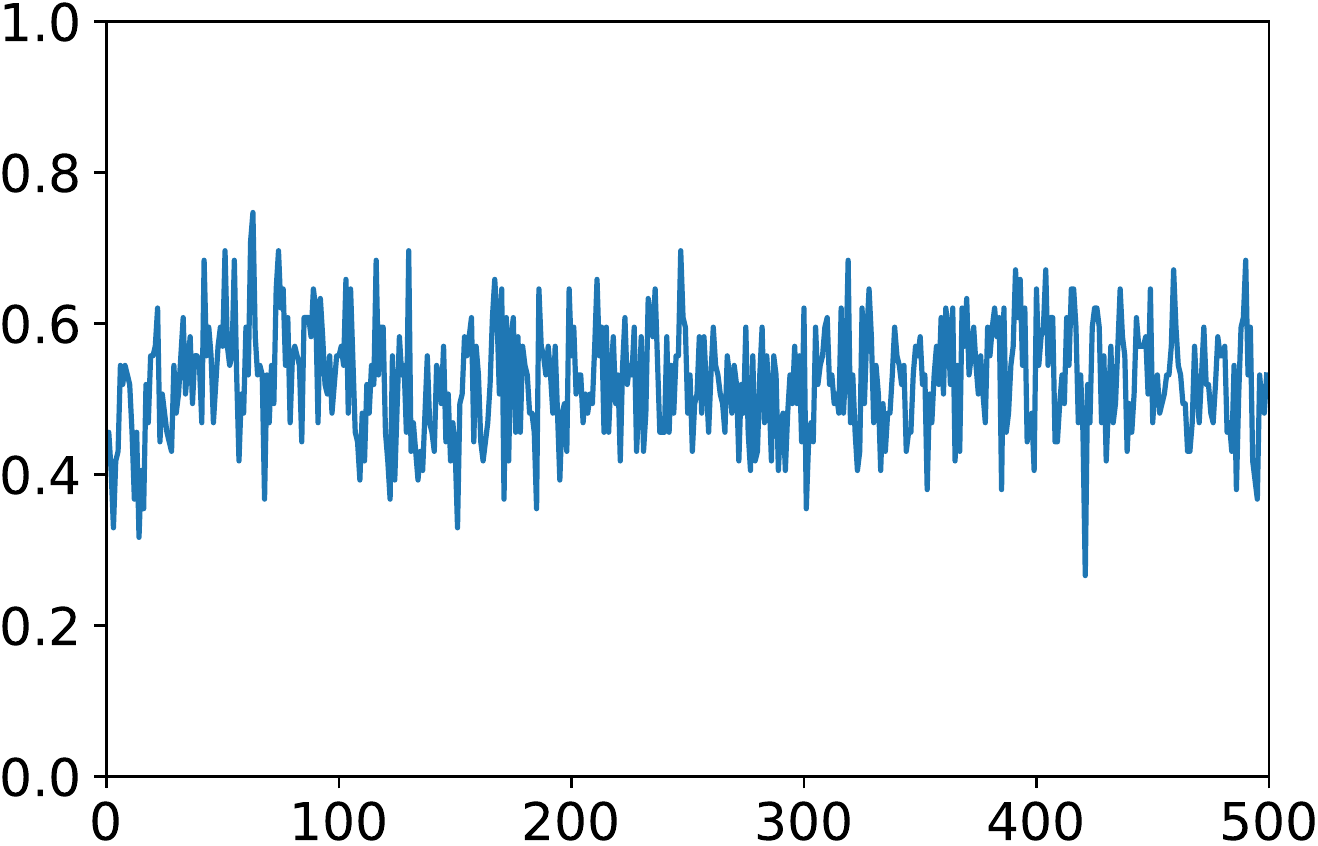}}
    \hspace{5pt}
    \subfigure[the $1000^{th}$ window]{
        \includegraphics[width=4cm]{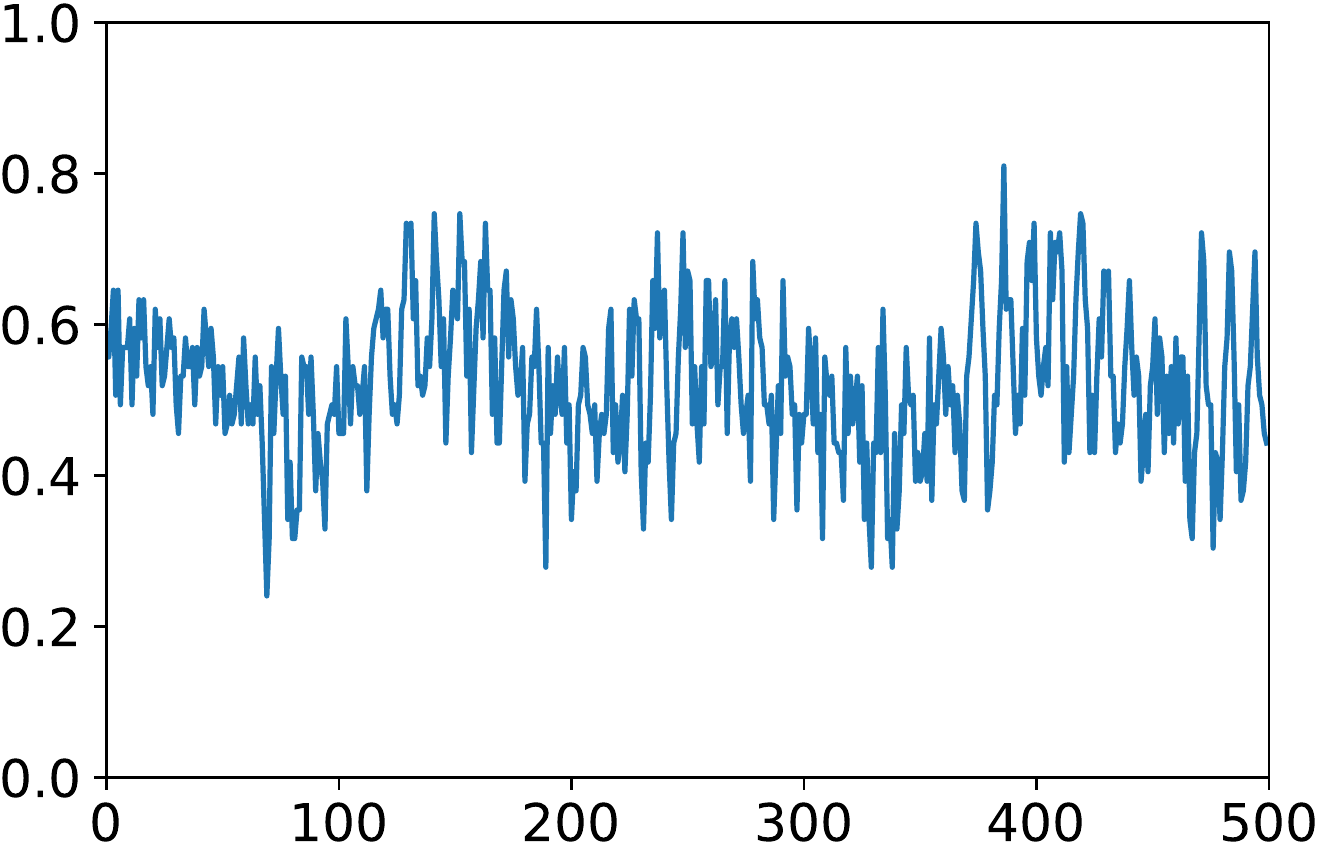}}
    \hspace{5pt}
    \subfigure[the $2000^{th}$ window]{
        \includegraphics[width=4cm]{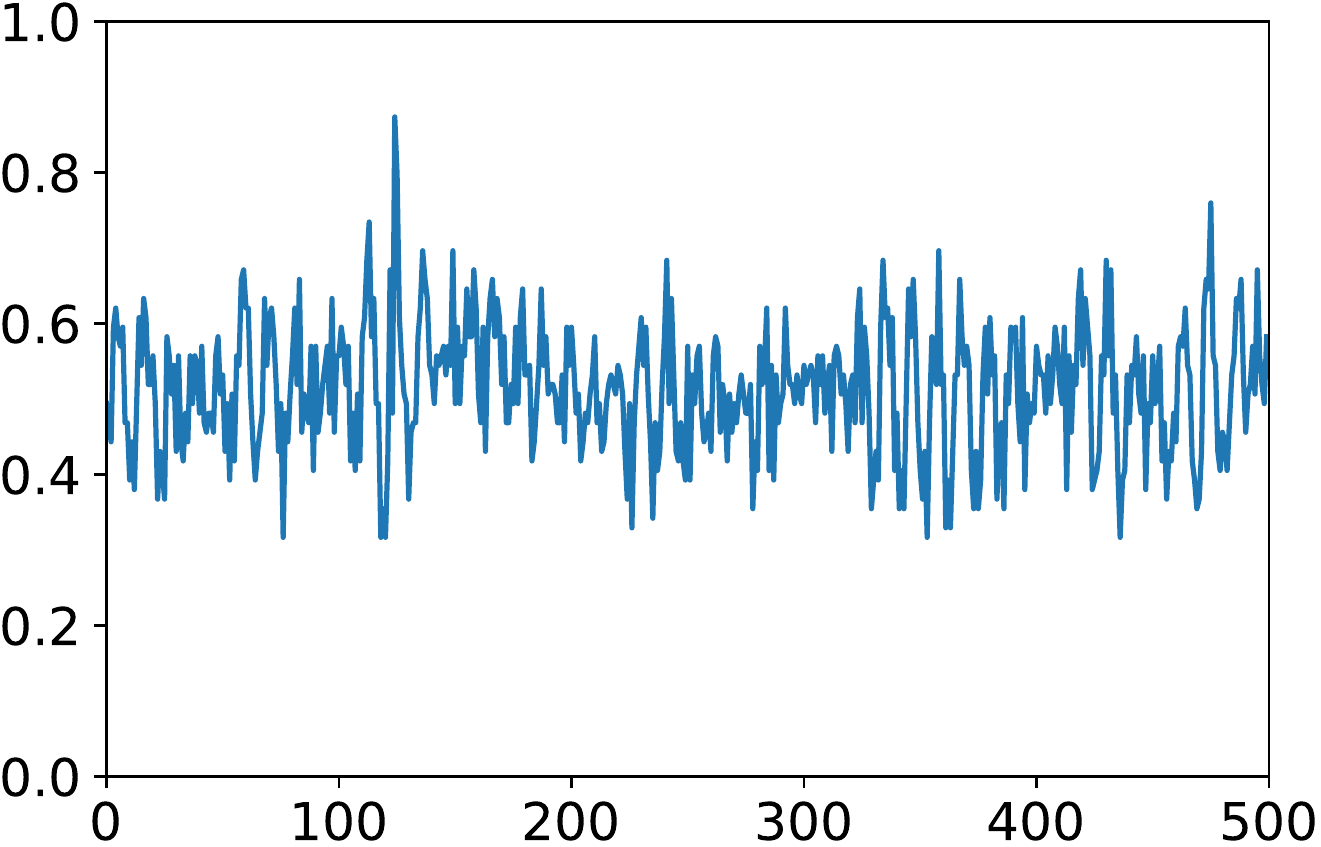}}
    \hspace{5pt}
    \subfigure[the $3000^{th}$ window]{
        \includegraphics[width=4cm]{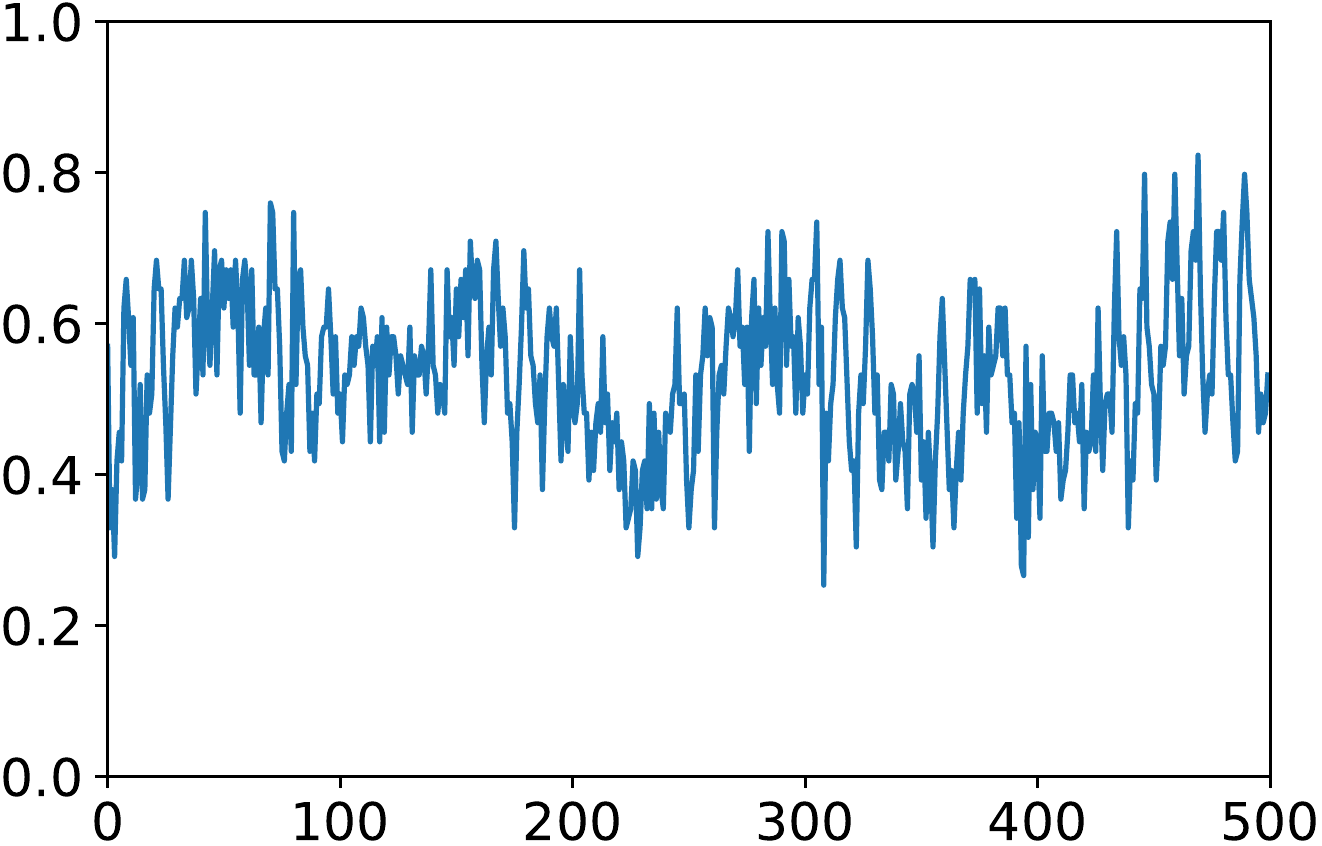}}
    \hspace{5pt}
    \subfigure[the $4000^{th}$ window]{
        \includegraphics[width=4cm]{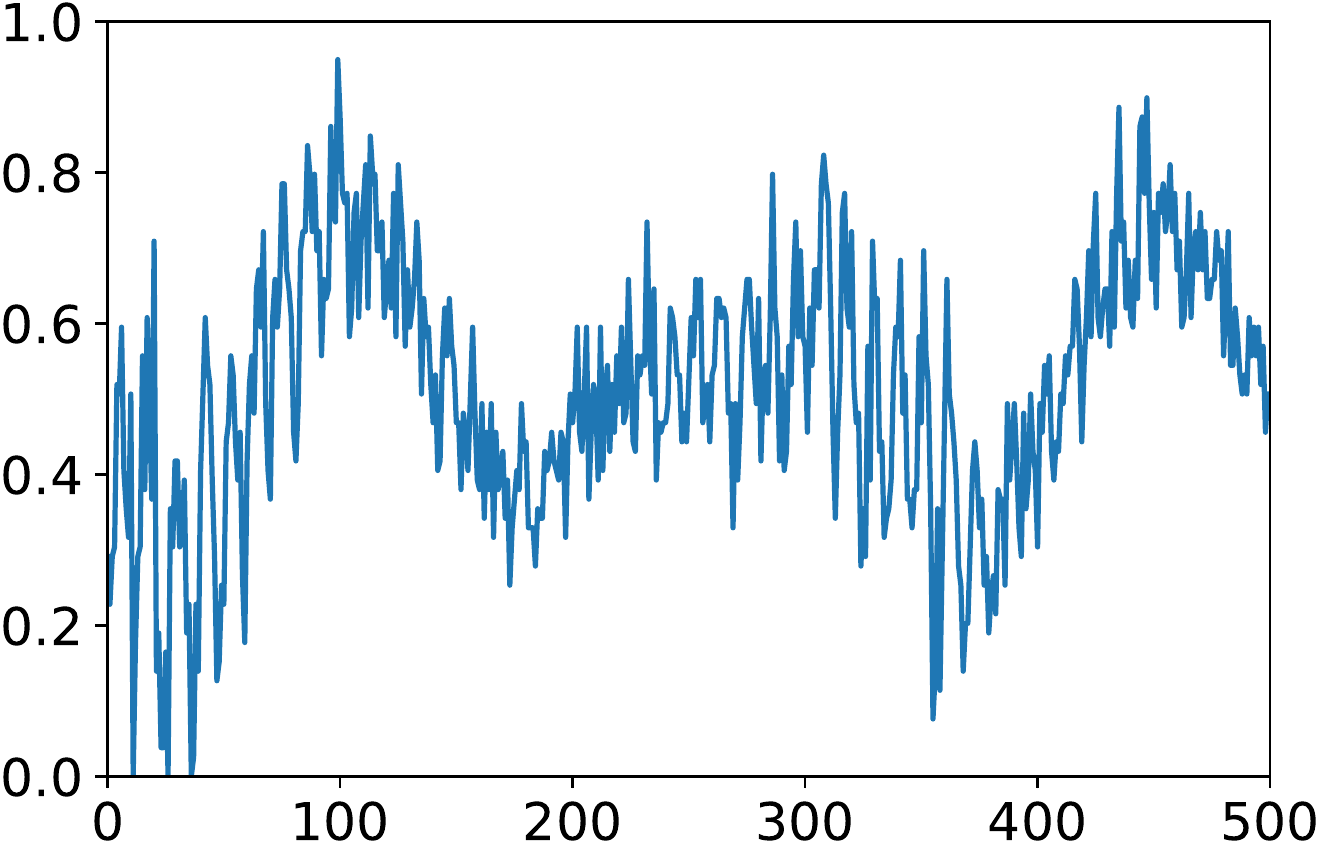}}
    \hspace{5pt}
    \subfigure[the $5000^{th}$ window]{
        \includegraphics[width=4cm]{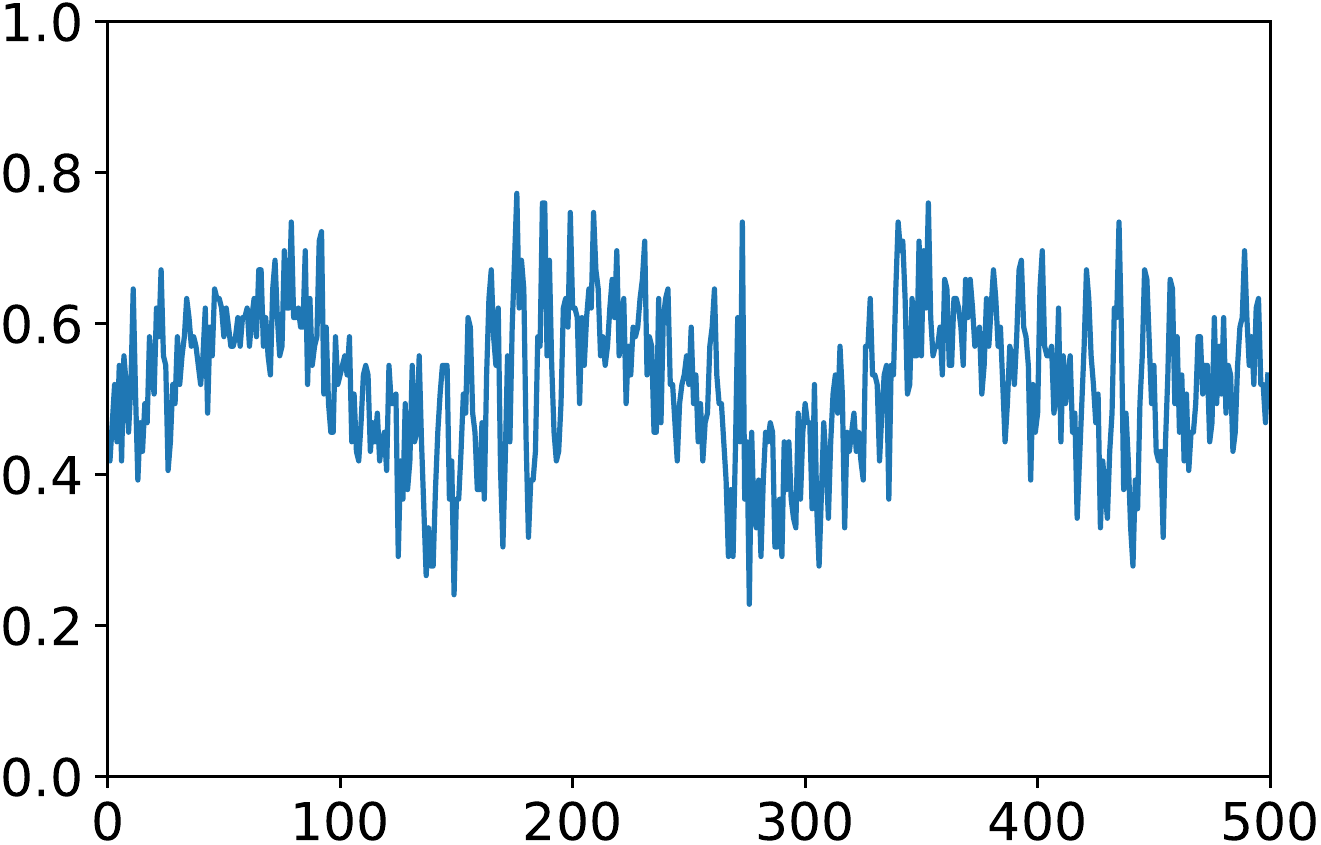}}
    \caption{Extracted normal samples from the first inspection of MP1}
    \label{windows_81}
\end{figure*}

We evaluated the predicted results on multiple testing time series from inspections with labeled anomalies. As is shown in Table \ref{tab:defect_count}, inspection 3 and 5 are clean inspections that contain no labeled anomalies. They are taken as training and validation sets, respectively. Inspections 1, 2 and 4 are employed as testing sets. A number of extracted subsequences from the first milepost are shown in Fig.~\ref{windows_81}. In this figure, the samples are all extracted from inspection 3, which has no reported anomalies. As shown, the patterns of the selected six samples are different from each other. This is fairly understandable given the diversity and variety of the real-world environment. 

This dataset has been used to evaluate the DEGAN framework, as well as some of its elements. To this end, we have used the framework to assess (1) how GAN hyperparameters affect its training; (2) how sliding window size affects the performance; (3) how clustering affects the performance; (4) how different GAN architectures affect the performance; and (5) how the model selection using unlabeled data is compared to the case of using labeled data. How data from different inspections and mileposts were used in these analyses have been summarized in Table \ref{tab:data_task}. 

\begin{table}[]
\centering
\caption{Explanation of dataset usage on different tasks}
\label{tab:data_task}
\begin{tabular}{@{}lll@{}}
\toprule
Evaluation Task                                    & Inspection                                                                                                                                          & MP                             \\ \midrule
Impact of GAN hyperparameters           & Train on 3 | Validate on 5                                                                                                                           & e.g. 4                          \\
Performance with different window lengths & \multirow{3}{*}{Train on 3 | Validate on 5 | Test on 1, 2, 4}                                                                                           & \multirow{3}{*}{1, 2, 3, 4, 5} \\
Impact of clustering                    &                                                                                                                                                     &                                \\
Impact of GAN achitectures                        &                                                                                                                                                     &                                \\
$\mathscr{D}$ model validation/selection          & \begin{tabular}[l]{@{}l@{}}Supervised: Train on 3 | Validate on 2 | Test on 1, 4\\ Unsupervised: Train on 3 | Validate on 5 | Test on 1, 4\end{tabular} & 1, 2, 3, 4, 5                  \\ \bottomrule
\end{tabular}
\end{table}

\subsection{Evaluation metrics}
\label{sec: Evaluation metric presentation}



Examples of predicted anomalies, real defects and probability distribution functions from KDE for each milepost with a window size of 200 are shown in Fig.~\ref{test_results}. To facilitate the visualization, the vertical acceleration and the probability distribution function were both normalized to [0,1]. The real anomalies are marked as red crosses on the acceleration time series, and the predicted ones (middle point of corresponding sliding windows) are depicted by green dots. Predicted anomalies on the KDE-generated probability distribution function are marked as blue dots in Fig.~\ref{test_results}. From this plot, we can see the selected $\mathscr{D}$ model has a good performance, where the predicted anomalies are in the vicinity of certain real defects. Therefore, to quantify the overall performance, a vicinity range (i.e., tolerance) should be considered, which is denoted as $r_t$ in Eq. (\ref{TP_FN_FP}). This tolerance $r_t$ shows how effective the algorithm is in detecting anomalies in a given vicinity of the real anomalies. 




In this case study, we explicitly define True Positives (TP), False Negatives (FN) and False Positives (FP) as Eq. (\ref{TP_FN_FP}), where we denote the location of a labeled ground truth anomaly and its closest anomaly prediction as $d$ and $p_{closest}$. We considered the ground truth anomalies that have a predicted anomaly in their vicinity as TP, while others that do not have a predicted anomaly in their vicinity as FN. FP is defined as an anomaly prediction that does not have a real defect in its vicinity, where $p$ and $d_{closest}$ represent the location of a predicted anomaly and its closest real defect, respectively. Thus, recall and precision can be calculated according to Eq. (\ref{Recall}) and Eq. (\ref{Precision}). 

\begin{equation}
\begin{cases}TP & \mid d-p_{closest}\mid\leq r_t\\FN & \mid d-p_{closest}\mid > r_t\\FP & \mid p-d_{closest}\mid > r_t \end{cases}
\label{TP_FN_FP}
\end{equation}

\begin{equation}
Recall = \frac{TP}{TP+FN}
\label{Recall}
\end{equation}

\begin{equation}
Precision = \frac{TP}{TP+FP}
\label{Precision}
\end{equation}

To clarify the calculation of evaluation metrics, we have provided an example as follows. Fig.~\ref{test_results} shows the anomaly detection result for the second inspection of MP3. There are five labeled anomalies (i.e., \(d_1\) to \(d_5\)) located at 15, 350, 351, 2710, 2711, and two anomaly predictions \(p_1\) and \(p_2\) located at 398, 2759 feet from left to right. Assuming the acceptable vicinity threshold of 100 feet ($\approx$30 m), then considering the real defects, we observe that \(d_2\), \(d_3\), \(d_4\)  and  \(d_5\)  were successfully detected, while \(d_1\) was missed. For the predictions, none of them is an FP. Hence, in this example, we have: recall $ = \frac{TP}{TP+FN} = \frac{4}{4+1} = 80\%$, precision $ = \frac{TP}{TP+FP} = \frac{4}{4+0} = 100\%$.


\subsection{Results and analysis}
The case study was empowered by GPU (NVIDIA GeForce RTX2080 Ti), with the aid of Cuda (v11.1.0), Cudnn and TensorFlow-gpu. In this section, we have presented the results for the evaluation of the proposed framework including comparative assessments by considering alternative approaches to some of the elements of the framework. We have further evaluated the influence of hyperparameters, sliding window sizes, time series clustering, alternative GAN architectures, and compared the supervised/unsupervised model validation methods.

\subsubsection{Discussion of GAN hyperparameter effects on training}
\begin{figure*}[h]
  \centering
  \includegraphics[width=7cm]{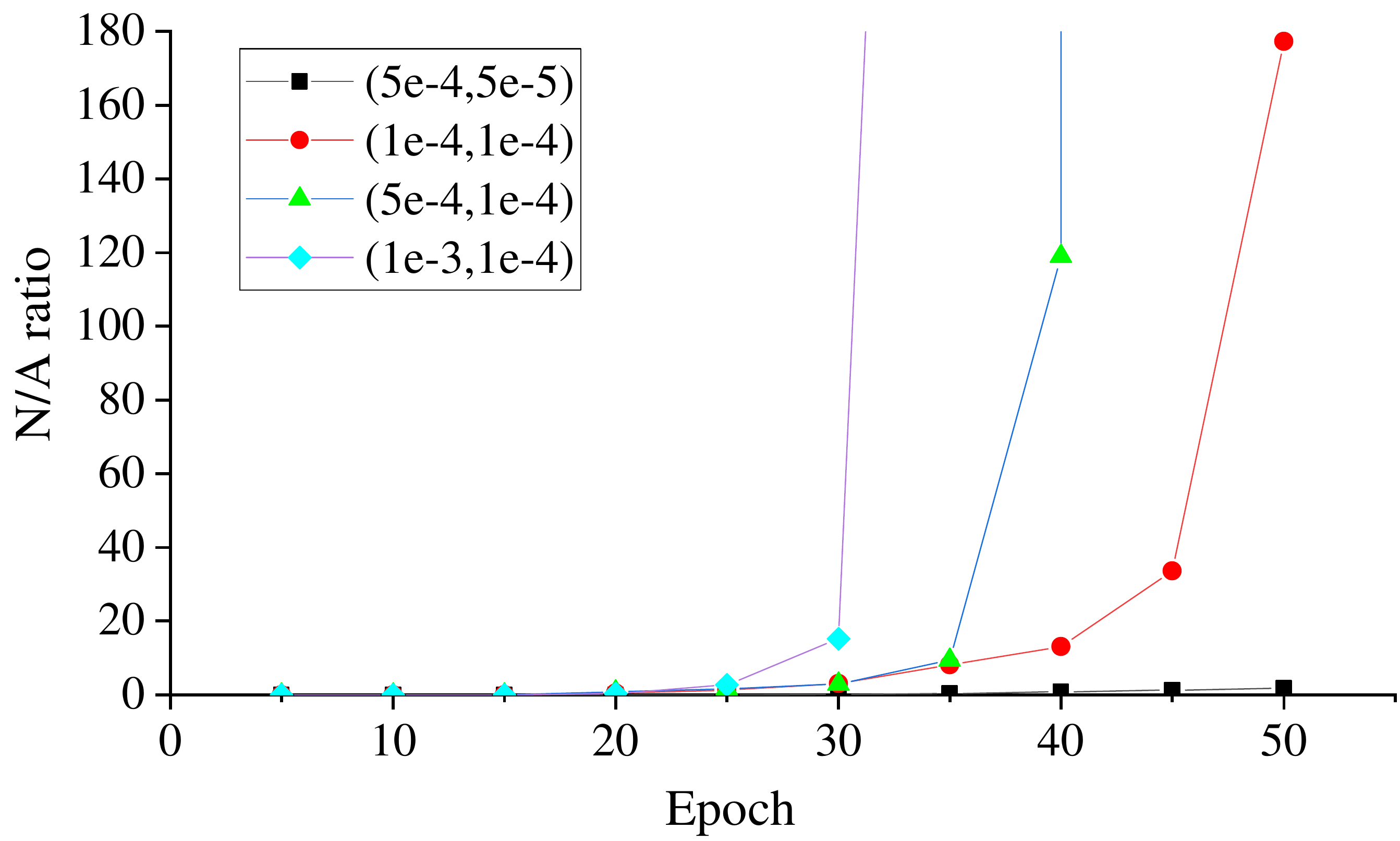}
  \caption{Example of $N/A$ ratios for different learning rates - trained on MP4 of inspection 3 and evaluated on inspection 5. }
  \label{fig:lr}
\end{figure*}

\
\label{sec: Discussion on hyperparameter configuration}
Similar to all deep learning problems, learning rates have a considerable influence on GAN training. To show the impact, we have demonstrated these effects with $N/A$ ratios mentioned in Section \ref{sec:Discriminator model selection}. Four models with different $\mathscr{G}$'s and $\mathscr{D}$'s learning rate pairs $(g\_lr, d\_lr)$ were first trained on inspection 3 and then evaluated on inspection 5 on MP4. The models were trained for 50 epochs and the $\mathscr{D}$ was saved every 5 epochs. The learning rate pairs and the $N/A$ ratios obtained from inspection 5 are shown in Fig.~\ref{fig:lr}. We can see that larger learning rates lead to an earlier and quicker rise in the $N/A$ ratio curve. However, a too large learning rate makes it difficult to identify the best epoch as $N/A$ curve jumps from overestimating to underestimating anomalies. On the other hand, a too small learning rate adds to the computational cost and makes too many false positive anomaly predictions. Therefore, in our case study, we first conducted a grid search on 10 pairs of learning rates with $g\_lr$ = [1e-4, 5e-4, 1e-3, 2e-3, 6e-3] and $d\_lr$ = [1e-4, 5e-4] to create various $N/A$ curves. Then, we selected the best learning rate pair which provided the highest but not infinite $N/A$ ratio at epoch 50 as discussed in Section 3.2. Moreover, we can also see from this example that the training epoch is another significant factor. How/whether we conduct a  $\mathscr{D}$ model selection (i.e., early stop of training) with labeled/ unlabeled data is further discussed in Section \ref{sec: Impact of labeled vs. unlabeled Discriminator validation} 

\subsubsection{DEGAN performance using different window lengths}\
\label{sec: DEGAN performance using different window sizes}

\begin{figure*}[h]
  \centering
  \includegraphics[width=13cm]{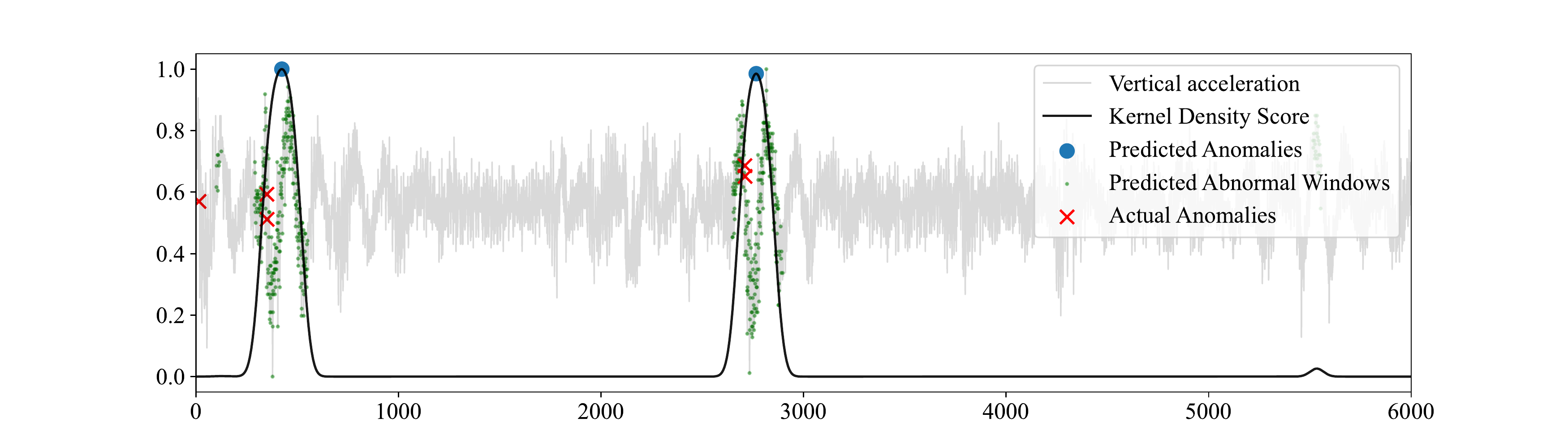}
  \caption{Examples of anomaly detection results for MP3}
  \label{test_results}
\end{figure*}


   
   
   
   
  
The overall performance of DEGAN was evaluated based on all the mileposts. The performance scores - recall, precision and F1 - with different \(r_t\) values are shown in Table \ref{tab:wl_evaluation}. Five different sliding window sizes (50, 100, 200, 300 and 500) were compared. Examples of MP3 with a window size of 200 are shown in Fig.~\ref{test_results}. Considering the randomness introduced by random noise input and random initialization for GAN networks, the predictions were not the same in different runs. Therefore, we have presented evaluation scores as the average performance of three runs in Table \ref{tab:wl_evaluation}. A higher value of \(r_t\) in the vicinity of real defects means a higher degree of fault tolerance. Therefore, the highest evaluation scores in Table \ref{tab:wl_evaluation} were obtained when \(r_t\) is 200 feet ($\approx$60 m), while smaller ranges can result in a stricter location identification in field evaluations. The highest average recall is 0.88, which was obtained with a window size of 200. The highest average precision and F1 were 0.86 and 0.83, respectively, when using a window size of 100. Although a higher recall for anomaly detection is of great importance to ensure that false negatives are minimized, the proposed framework also shows a promising performance in resulting in a high precision which helps reduce assessments of the physical system (i.e., field assessments). 
\begin{table}[]
\centering
\caption{Evaluation scores of different window lengths ($wl$) under tolerance ranges ($r_t$)}
\label{tab:wl_evaluation}
\begin{tabular}{@{}cccccccccc@{}}
\toprule
$r_t$ (feet) & \multicolumn{3}{c}{100}   & \multicolumn{3}{c}{150}   & \multicolumn{3}{c}{200}                       \\ \midrule
$wl$  & Recall & Precision & F1   & Recall & Precision & F1   & Recall        & Precision     & F1            \\ \midrule
50         & 0.70   & 0.76      & 0.73 & 0.73   & 0.79      & 0.76 & 0.76          & 0.82          & 0.79          \\
100        & 0.71   & 0.77      & 0.74 & 0.77   & 0.82      & 0.79 & 0.80          & \textbf{0.86} & \textbf{0.83} \\
200        & 0.80   & 0.66      & 0.72 & 0.83   & 0.67      & 0.74 & \textbf{0.88} & 0.70          & 0.78          \\
300        & 0.80   & 0.60      & 0.69 & 0.83   & 0.61      & 0.70 & 0.85          & 0.63          & 0.72          \\
500        & 0.59   & 0.41      & 0.49 & 0.84   & 0.57      & 0.68 & 0.85          & 0.58          & 0.69          \\ \bottomrule
\end{tabular}
\end{table}


\subsubsection{Impact of clustering}
\label{sec: Influence of clustering}
\begin{figure*}
  \centering
        \includegraphics[width=3.2cm]{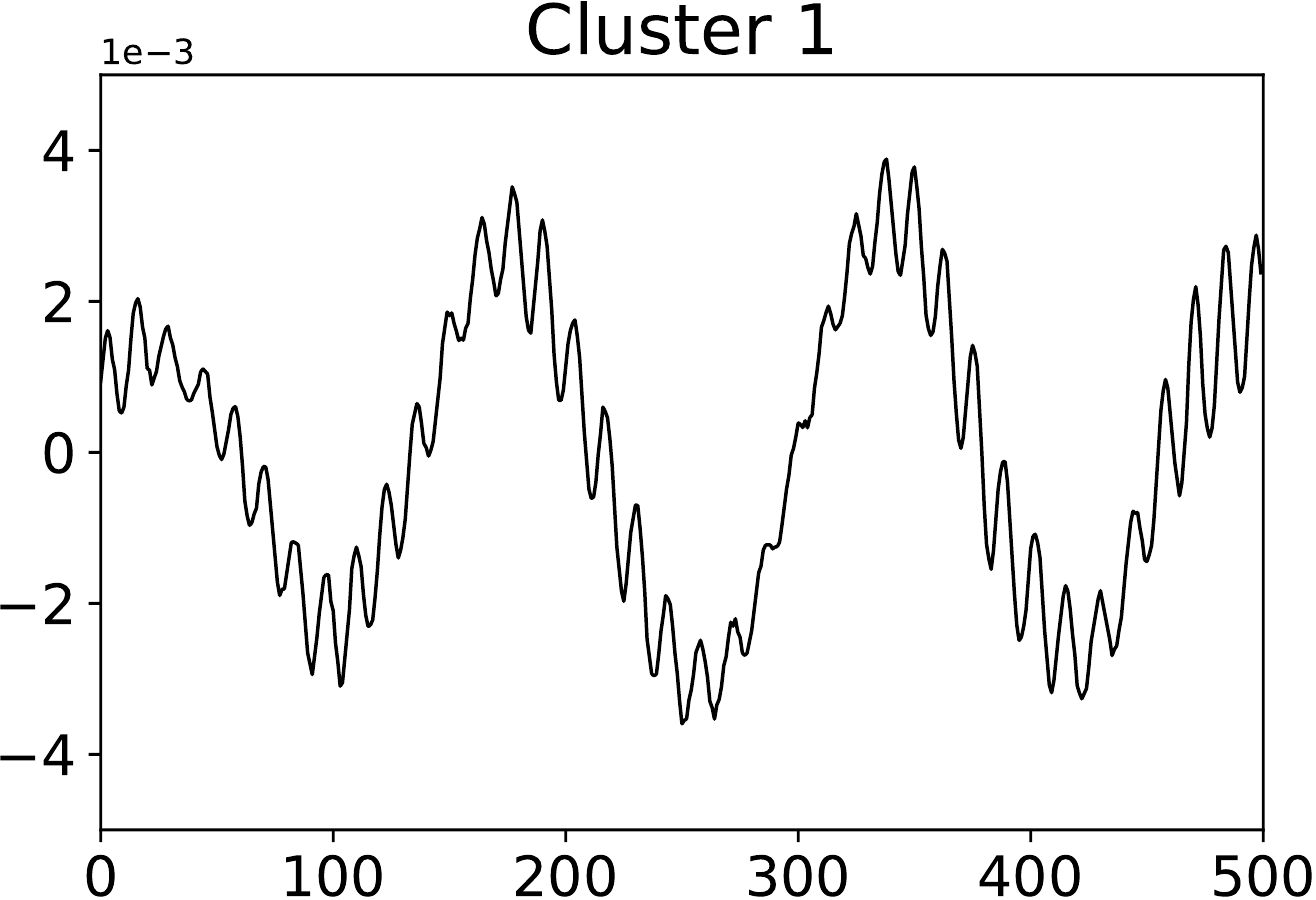}   
        \includegraphics[width=3.2cm]{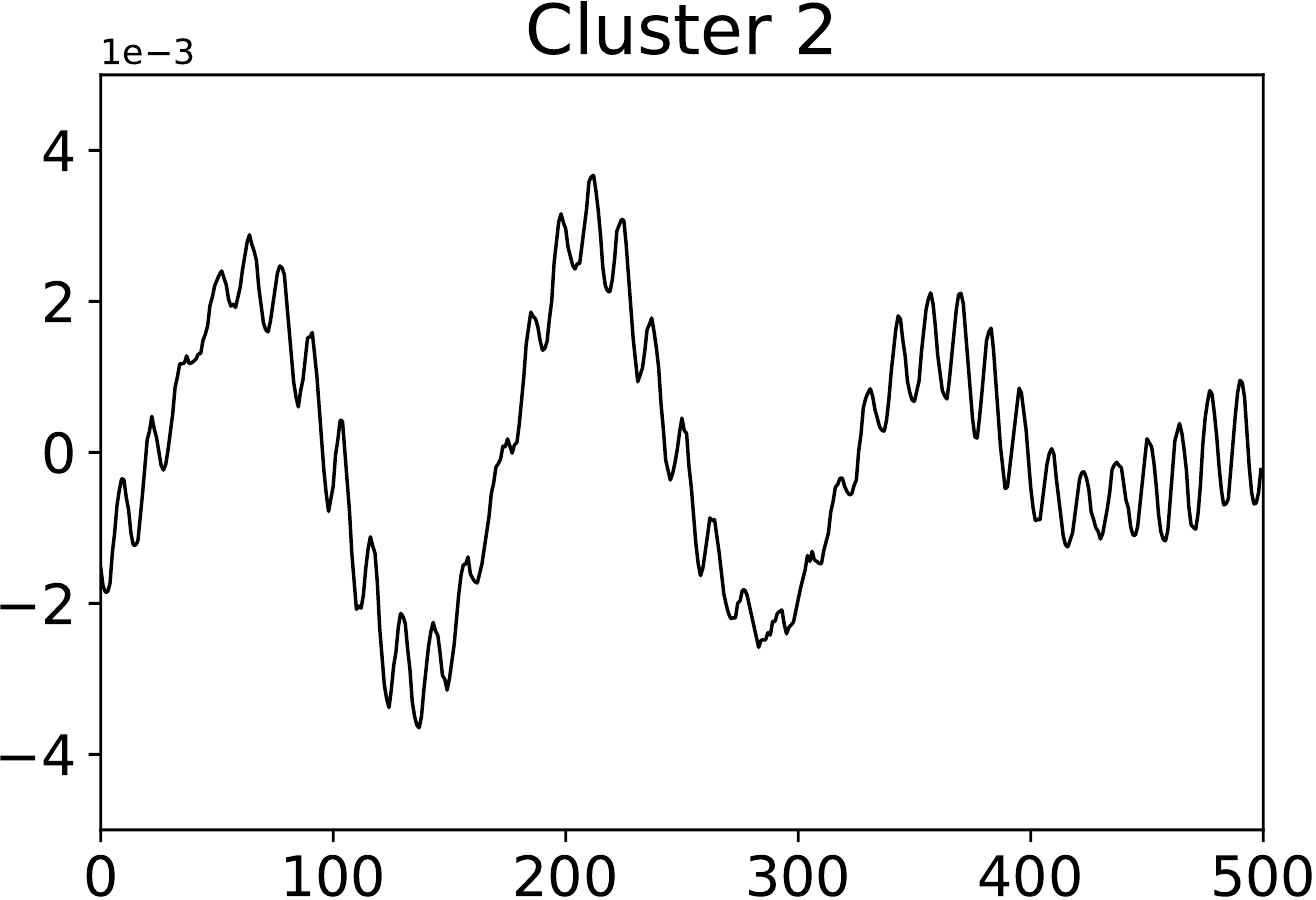}   
        \includegraphics[width=3.2cm]{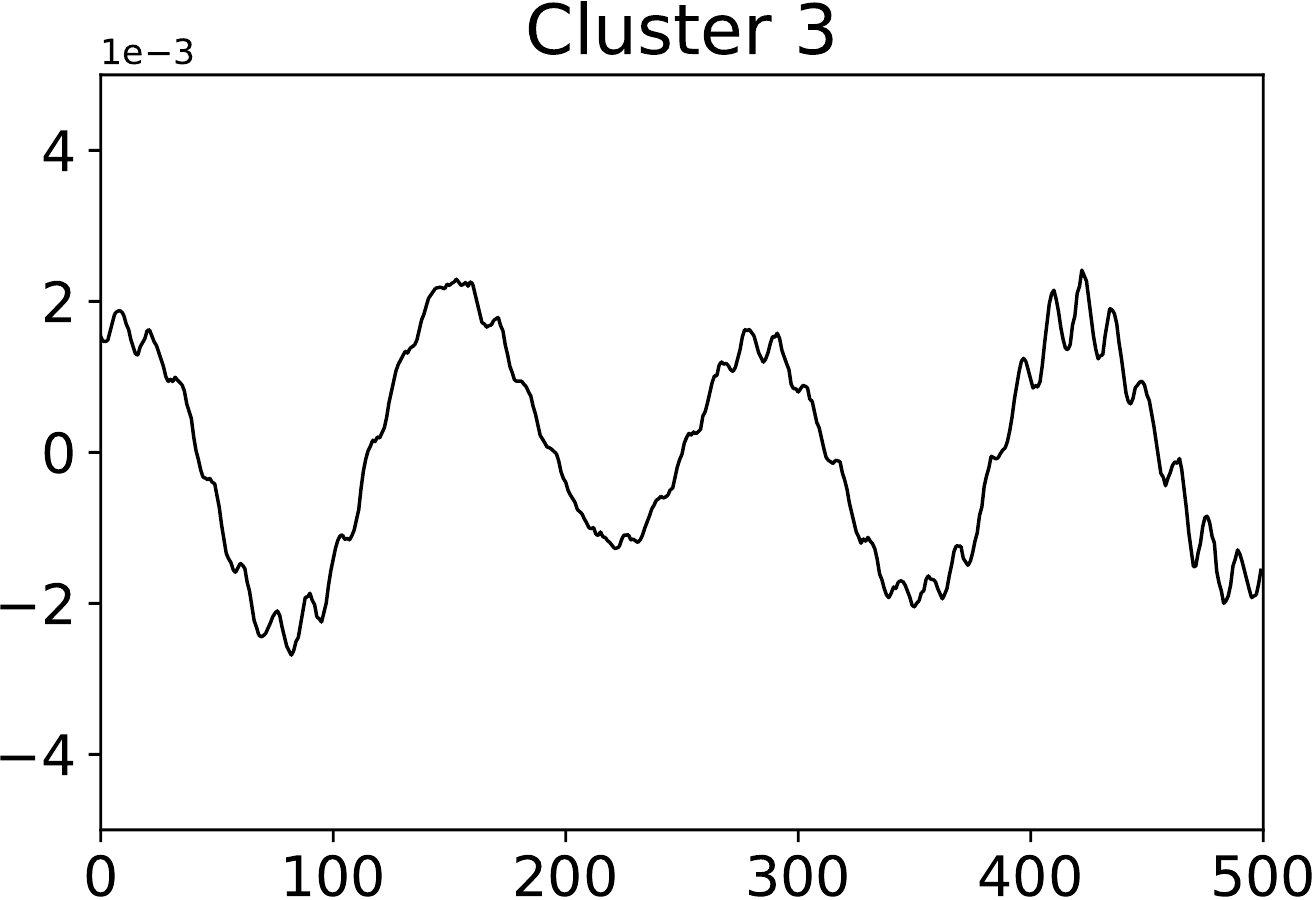}   
        \includegraphics[width=3.2cm]{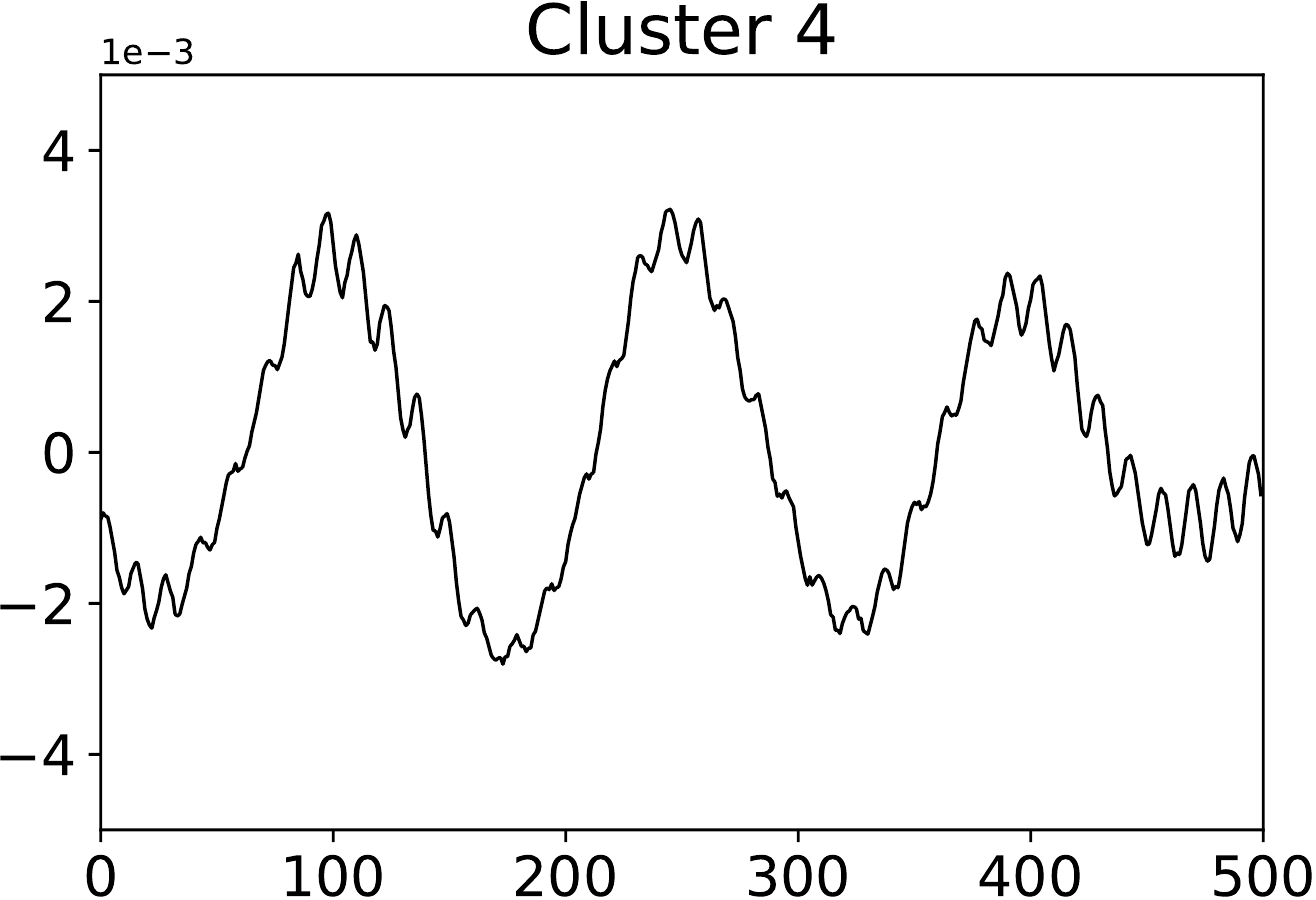}     
        \includegraphics[width=3.2cm]{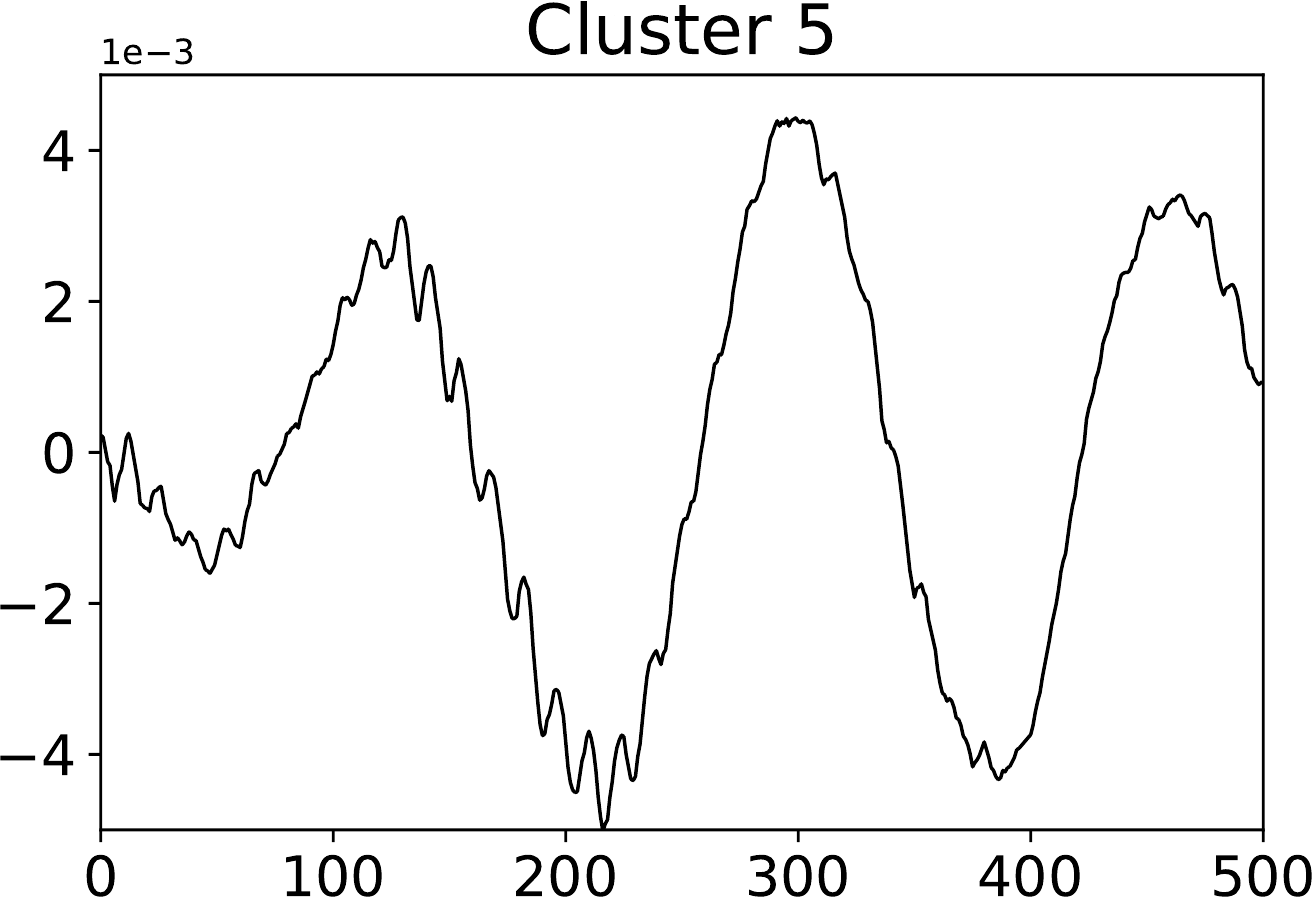}   
    \caption{Centroids of Euclidean clustering (k=5) on MP1}
    \label{centroids_euc}
\end{figure*}

\begin{figure*}
  \centering
        \includegraphics[width=3.2cm]{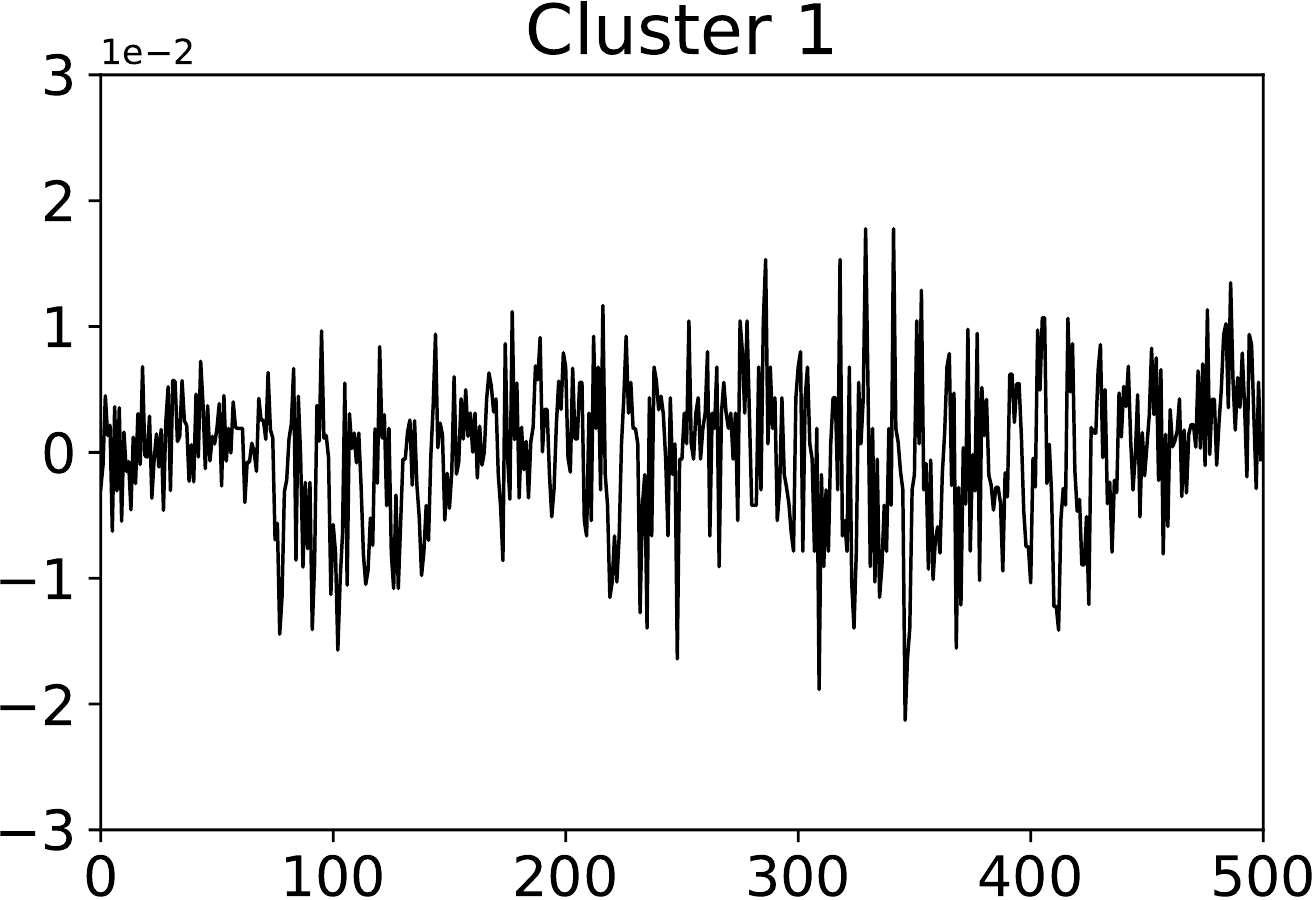}  
        \includegraphics[width=3.2cm]{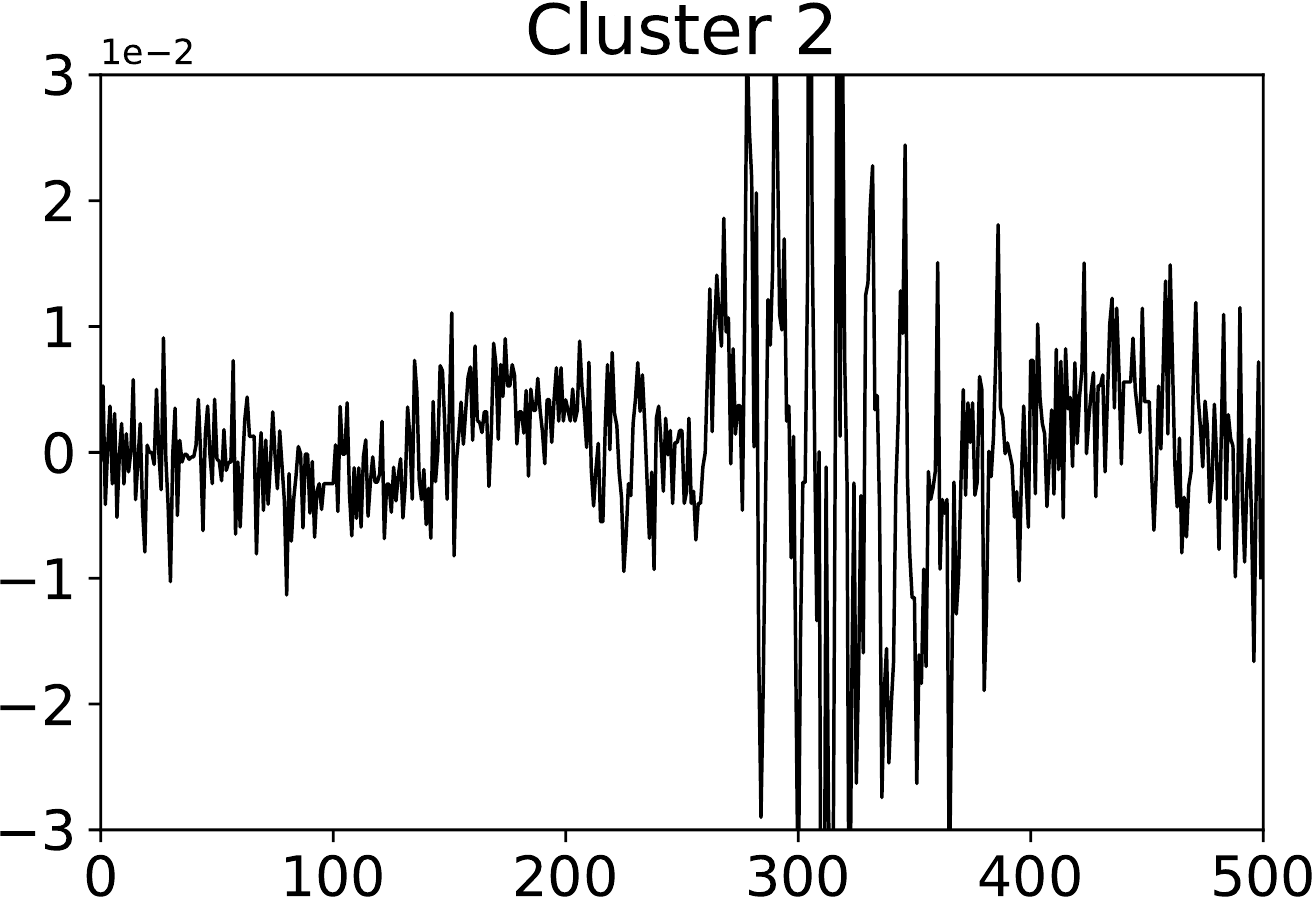}  
        \includegraphics[width=3.2cm]{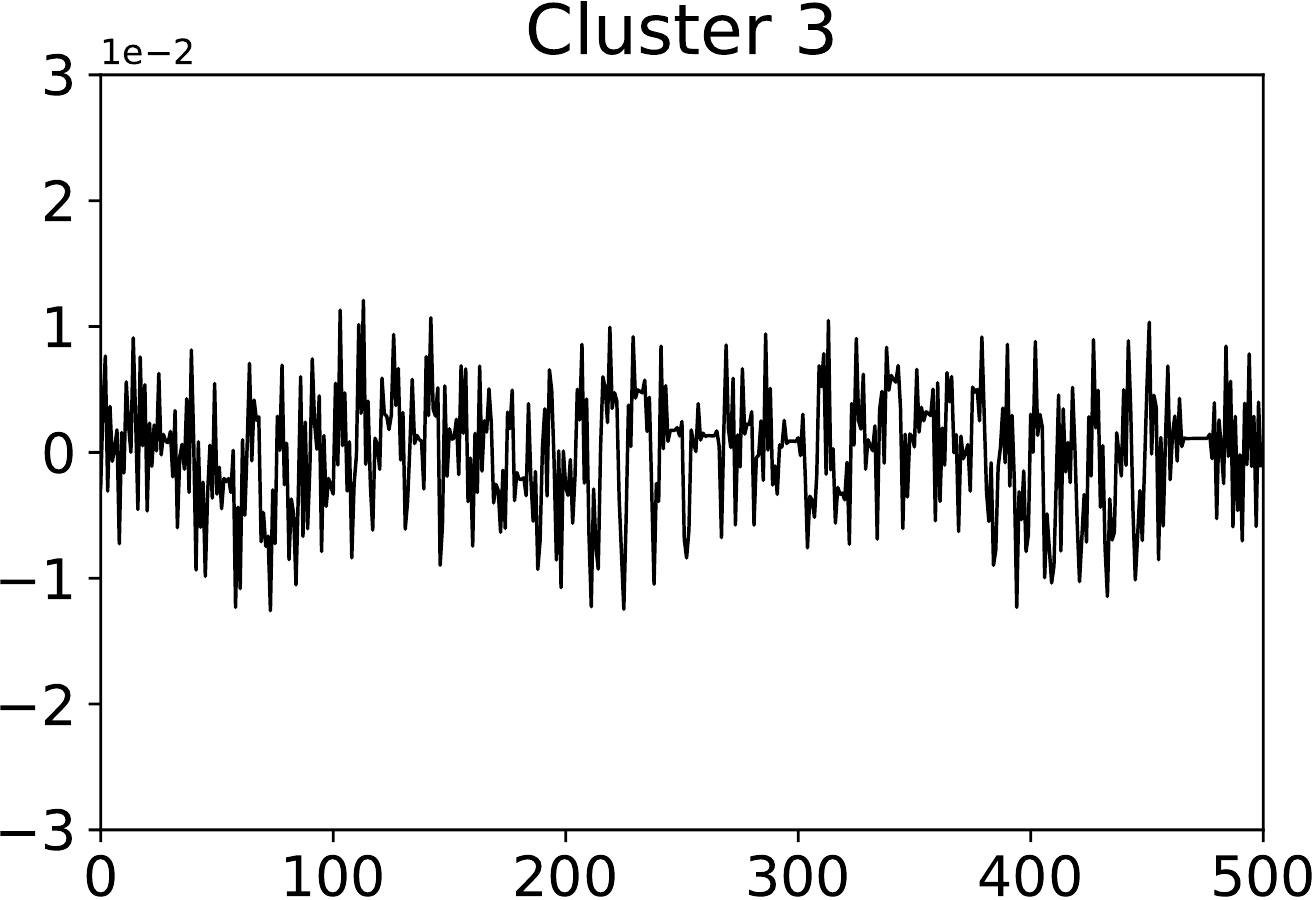}  
        \includegraphics[width=3.2cm]{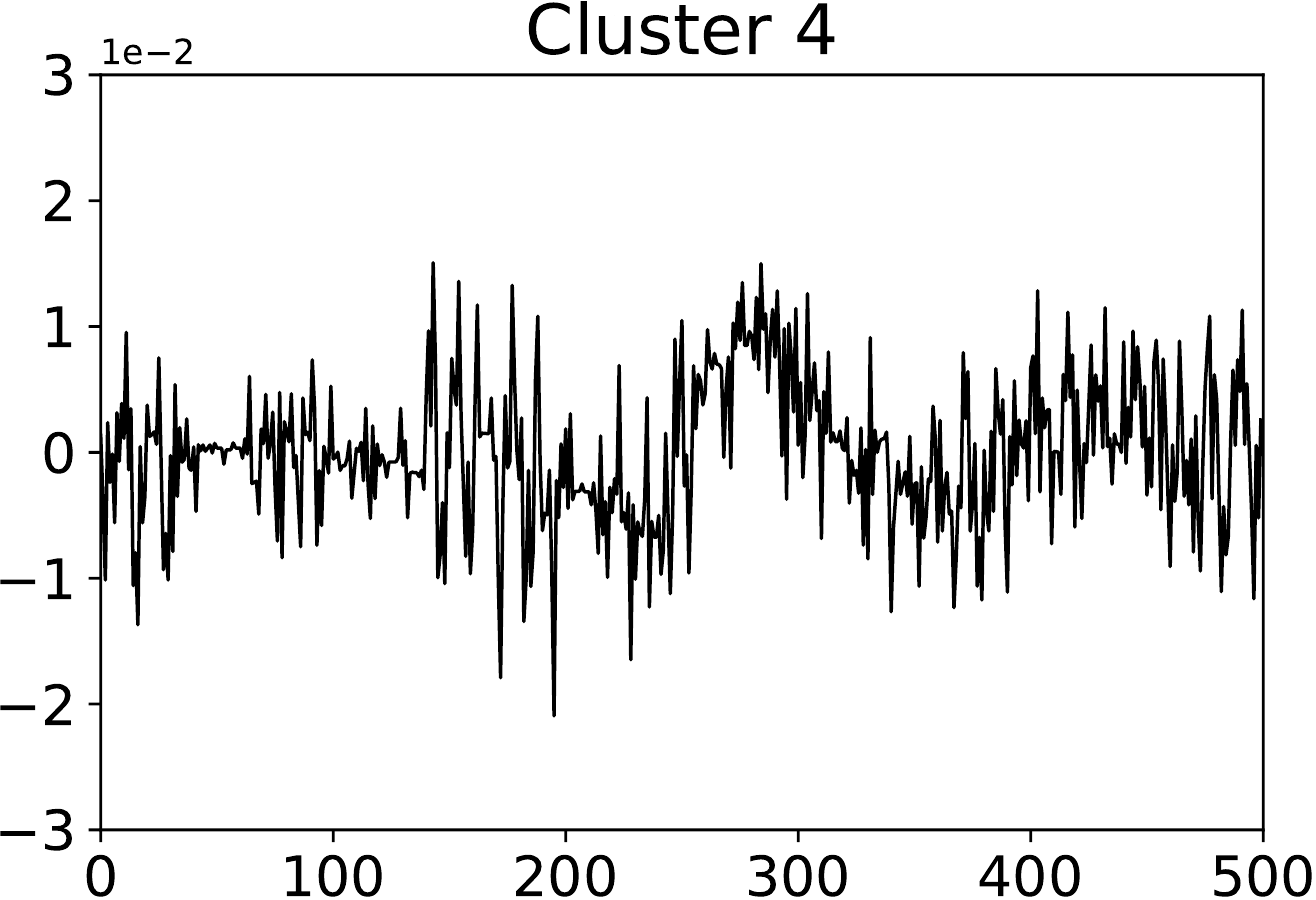}       
        \includegraphics[width=3.2cm]{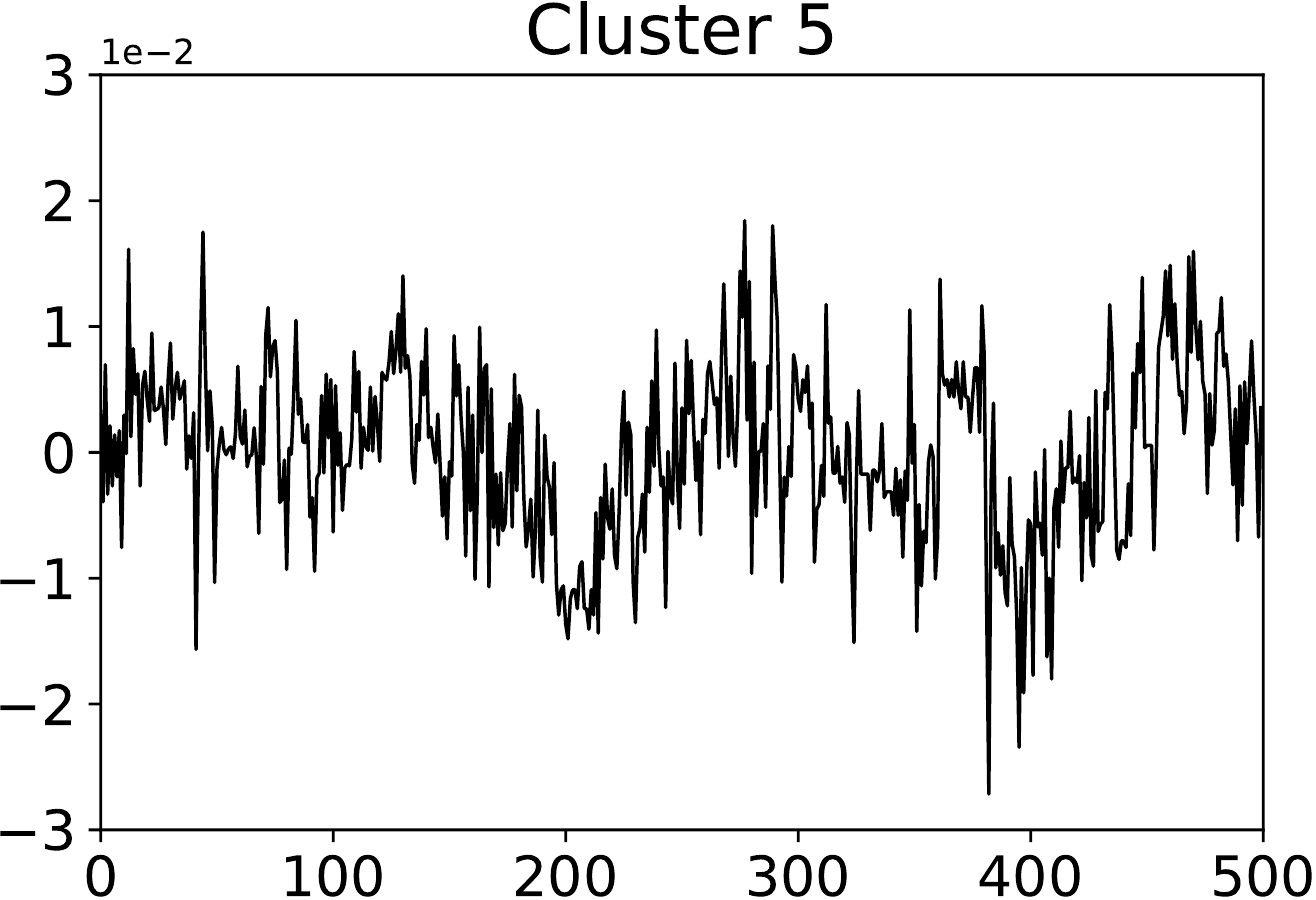}  
    \caption{Centroids of DTW clustering (k=5) on MP1}
    \label{centroids_dtw}
\end{figure*}

To evaluate the impact of clustering on the performance of the proposed framework, we experimented with the motif-based GAN approach as discussed in Section \ref{sec: methodology - Motif-based GAN training}. In doing so, we have used k-means clustering with two distance metrics. Common distance metrics for the classification and clustering of time series data include Euclidean and Dynamic Time Warping (DTW). The comparison of the Euclidean and DTW clustering results of window size 500 (as an example) are shown in Fig.~\ref{centroids_euc} and Fig.~\ref{centroids_dtw}. The evaluation scores are shown in  Table~\ref{tab:clustering_evaluation}. Unlike the result of the pre-experiment on sinusoidal waves (see Section \ref{sec: methodology - Motif-based GAN training}), in this case, although motif-based GAN could potentially improve the precision at the cost of recall for larger window sizes, such as 500, it generally decreases the performance for smaller window sizes (i.e., 100 and 200). This phenomenon might be associated with the fact that there is less variety of patterns when smaller subsequences of time series are used, resulting in fewer clusters that are not as effective. It could also be less effective given that clustering reduces the training data population for each GAN model.


\begin{table}[]
\centering
\caption{Comparison of clustering methods}
\label{tab:clustering_evaluation}
\begin{tabular}{@{}ccccccccccc@{}}
\toprule
\multicolumn{2}{c}{$r_t$ (feet)}           & \multicolumn{3}{c}{100}   & \multicolumn{3}{c}{150}   & \multicolumn{3}{c}{200}                       \\ \midrule
$wl$            & Clustering method & Recall & Precision & F1   & Recall & Precision & F1   & Recall        & Precision     & F1            \\ \midrule
\multirow{3}{*}{100} & None              & 0.71   & 0.77      & 0.74 & 0.77   & 0.82      & 0.79 & \textbf{0.80} & \textbf{0.86} & \textbf{0.83} \\
                     & Euclidean         & 0.69   & 0.69      & 0.69 & 0.75   & 0.76      & 0.76 & 0.78          & 0.79          & 0.78          \\
                     & DTW               & 0.59   & 0.65      & 0.62 & 0.66   & 0.70      & 0.68 & 0.67          & 0.72          & 0.69          \\ \midrule
\multirow{3}{*}{200} & None              & 0.80   & 0.66      & 0.72 & 0.83   & 0.67      & 0.74 & \textbf{0.88} & \textbf{0.70} & \textbf{0.78} \\
                     & Euclidean         & 0.71   & 0.63      & 0.66 & 0.76   & 0.65      & 0.70 & 0.77          & 0.67          & 0.72          \\
                     & DTW               & 0.77   & 0.62      & 0.69 & 0.81   & 0.65      & 0.72 & 0.86          & 0.67          & 0.76          \\ \midrule
\multirow{3}{*}{500} & None              & 0.59   & 0.41      & 0.49 & 0.84   & 0.57      & 0.68 & \textbf{0.85} & 0.58          & 0.69          \\
                     & Euclidean         & 0.49   & 0.43      & 0.46 & 0.61   & 0.51      & 0.55 & 0.83          & 0.65          & \textbf{0.73} \\
                     & DTW               & 0.43   & 0.42      & 0.42 & 0.64   & 0.58      & 0.61 & 0.76          & \textbf{0.67} & 0.71          \\ \bottomrule
\end{tabular}
\end{table}

\subsubsection{Impact of alternative GAN architectures}
\label{sec: Influence of GAN architectures}
\begin{figure}
  \centering
    \subfigure[Dense-$\mathscr{D}$]{              
        \includegraphics[height=4cm]{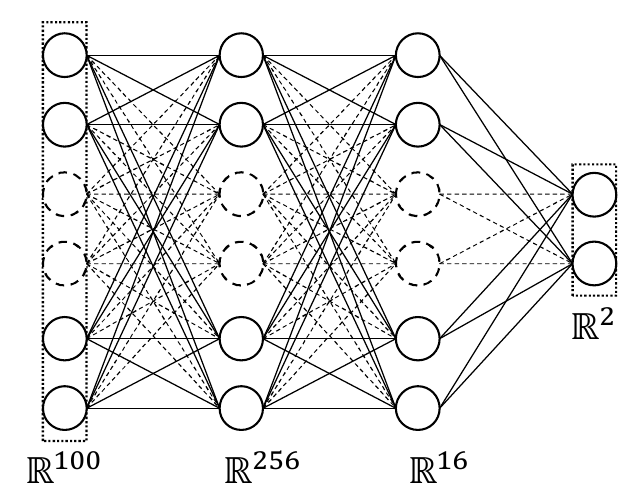}}
    \hspace{15pt}
    \subfigure[LSTM-$\mathscr{D}$]{
        \includegraphics[height=4cm]{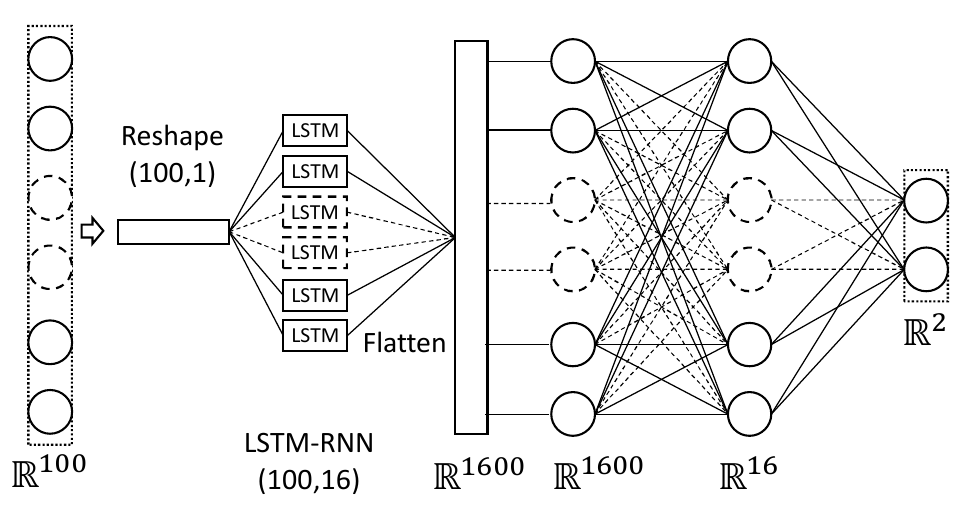}}
    \caption{Two alternative $\mathscr{D}$ architectures ($wl$=100)}
    \label{fig:alter_d_arch}
\end{figure}

In the context of anomaly detection for time series data, GAN networks usually employ Dense networks \cite{sun_time_2019}, CNN \cite{jiang2019}\cite{zhou2019beatgan} and RNN \cite{li_mad-gan_2019}\cite{geiger2020tadgan}\cite{bashar2020tanogan}\cite{niu2020lstm} as base models for $\mathscr{G}$ and $\mathscr{D}$. As we put an emphasis on the functionality of $\mathscr{D}$ in this study, aside from the base model CNN-$\mathscr{D}$ introduced in Section \ref{sec:The GAN architecture}, we compared two alternative $\mathscr{D}$ architectures for the $\mathscr{D}$ model, including a Dense architecture (Dense-$\mathscr{D}$) and an LSTM-RNN architecture (LSTM-$\mathscr{D}$). Details are shown in Fig. \ref{fig:alter_d_arch} and Table~\ref{tab:alter_d_arch}. The Dense-$\mathscr{D}$ simply stacks three Dense layers. The first two are activated by Tanh layers and the last one is followed by a Sigmoid layer with two output digits between 0 (real) and 1 (fake). The LSTM-$\mathscr{D}$ utilizes an LSTM-RNN layer with a neuron number of 16.

The anomaly detection results from these alternatives are shown in Table~\ref{tab:architecture_results}. CNN-$\mathscr{D}$ yielded the best performance among the three architectures and Dense-$\mathscr{D}$ also reached a promising score. Although LSTM-$\mathscr{D}$ may be more advantageous in cases when anomalies are strongly related to the context\cite{li_mad-gan_2019}\cite{geiger2020tadgan}\cite{bashar2020tanogan}\cite{niu2020lstm}, it did not show a good performance in our case study.

\begin{table}[]
\centering
\caption{Summary of two alternative $\mathscr{D}$ architectures ($wl$=100)}
\label{tab:alter_d_arch}
\begin{tabular}{@{}ccccccc@{}}
\multicolumn{1}{l}{} & (a) Dense-$\mathscr{D}$ & \multicolumn{1}{l}{} & \multicolumn{1}{l}{} & \multicolumn{1}{l}{} & (b) LSTM-$\mathscr{D}$ & \multicolumn{1}{l}{} \\ \cmidrule(r){1-3} \cmidrule(l){5-7} 
Layer                & Output Shape         & Parameter number              &  & Layer           & Output Shape   & Parameter number \\ \cmidrule(r){1-3} \cmidrule(l){5-7} 
InputLayer           & (None, 100)          & 0                    &  & InputLayer      & (None, 100)    & 0       \\
Dense (Tanh)         & (None, 256)          & 25856                &  & Reshape         & (None, 100,1)  & 0       \\
Dropout              & (None, 256)          & 0                    &  & LSTM-RNN        & (None, 100,16) & 1152    \\
Dense                & (None, 16)           & 4112                 &  & Dropout         & (None, 100,16) & 0       \\
Dense                & (None, 2)            & 34                   &  & Flatten         & (None, 1600)   & 0       \\
\multicolumn{1}{l}{} & \multicolumn{1}{l}{} & \multicolumn{1}{l}{} &  & Dense (Tanh)    & (None, 16)     & 25616   \\
\multicolumn{1}{l}{} & \multicolumn{1}{l}{} & \multicolumn{1}{l}{} &  & Dense (Sigmoid) & (None, 2)      & 34      \\ \cmidrule(r){1-3} \cmidrule(l){5-7} 
\end{tabular}
\end{table}



\begin{table}[]
\centering
\caption{Anomaly detection results of different $\mathscr{D}$ architectures}
\label{tab:architecture_results}
\begin{tabular}{@{}cccccccccc@{}}
\toprule
r\_t(feet)    & \multicolumn{3}{c}{100}                    & \multicolumn{3}{c}{150}                    & \multicolumn{3}{c}{200}                    \\ \midrule
$\mathscr{D}$ & Recall        & Precision     & F1            & Recall        & Precision     & F1            & Recall        & Precision     & F1            \\ \midrule
Dense-$\mathscr{D}$       & 0.65          & 0.74          & 0.69          & 0.68          & 0.77          & 0.72          & 0.70          & 0.79          & 0.74          \\
CNN-$\mathscr{D}$         & \textbf{0.71} & \textbf{0.77} & \textbf{0.74} & \textbf{0.77} & \textbf{0.82} & \textbf{0.79} & \textbf{0.80} & \textbf{0.86} & \textbf{0.83} \\
LSTM-$\mathscr{D}$        & 0.53          & 0.52          & 0.52          & 0.64          & 0.59          & 0.61          & 0.69          & 0.63          & 0.66          \\ \bottomrule
\end{tabular}
\end{table}

\subsubsection{Impact of labeled vs. unlabeled $\mathscr{D}$ validation}\
\label{sec: Impact of labeled vs. unlabeled Discriminator validation}
The significant impact of the number of training epochs on $\mathscr{D}$ model performance has been noted in Section \ref{sec: Discussion on hyperparameter configuration}. We have also introduced the necessity and methodology of the DEGAN model selection in Section \ref{sec:Discriminator model selection}. In order to demonstrate the feasibility and reliability of the proposed unsupervised method of $\mathscr{D}$ model selection, we have compared the model selection outcome based on clean time series data versus time series with labeled anomalies in this part. 

\begin{figure*}[h]
  \centering
  \includegraphics[width=6cm]{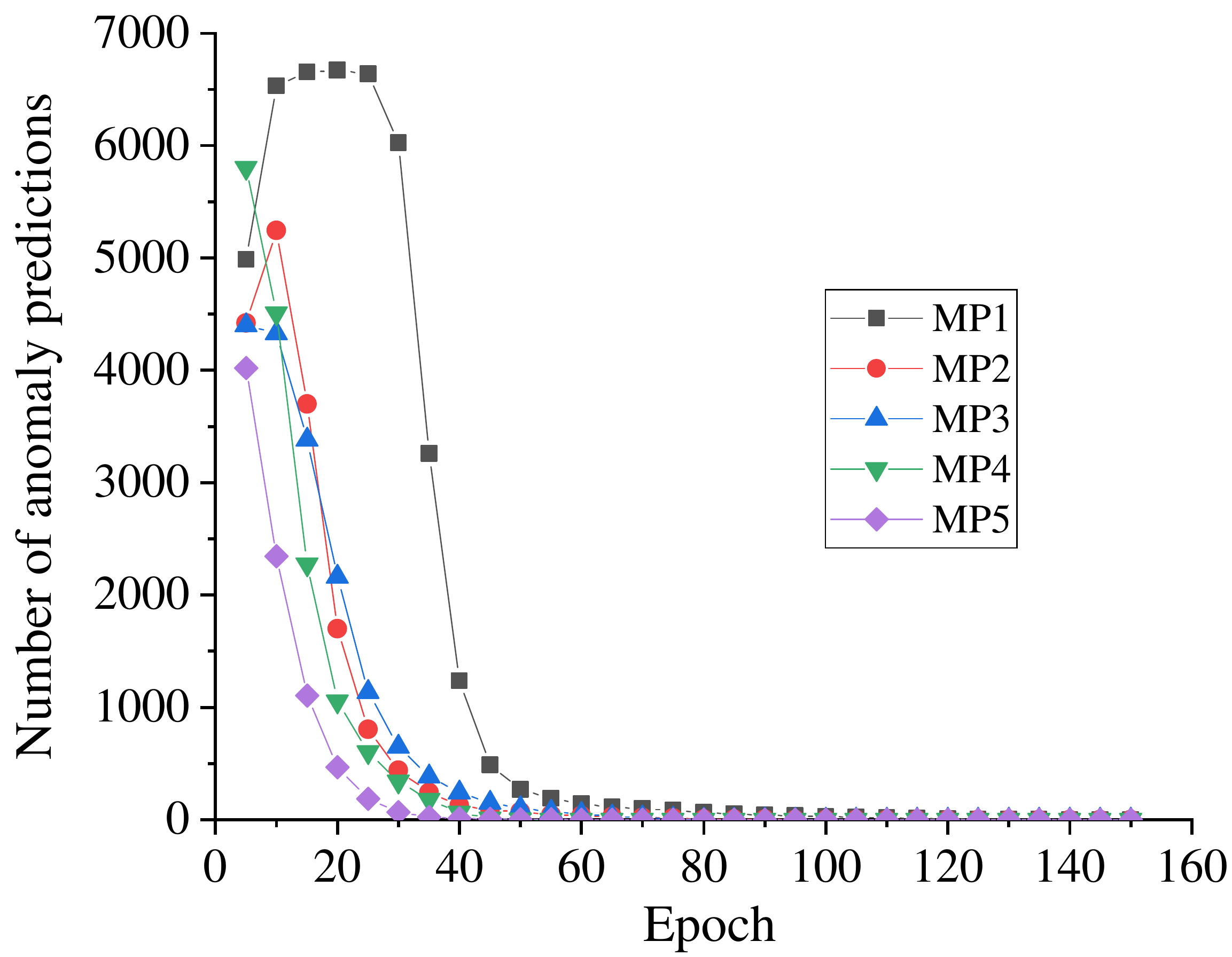}
  \caption{Anomaly prediction count on unlabeled validation data}
  \label{fig:aco_count_unlabeled}
\end{figure*}

\begin{figure*}[h]
  \centering
  \includegraphics[width=6cm]{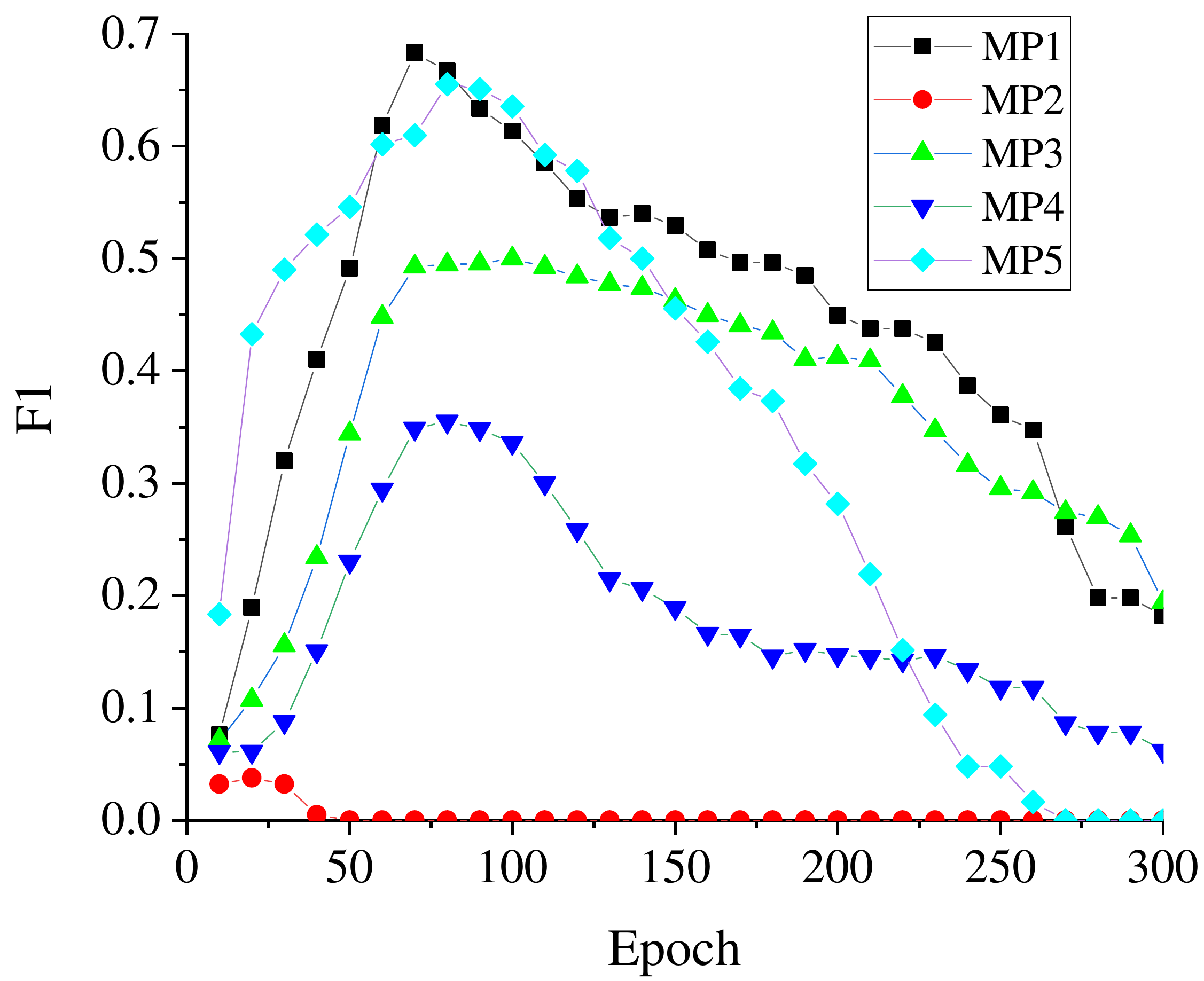}
  \caption{F1 score on labeled validation data}
  \label{fig:f1_labeled}
\end{figure*}

As shown in Table \ref{tab:data_task}, we trained two GAN models on inspection 3 (clean), and validated one of the $\mathscr{D}$ models on inspection 2 (labeled) and another one on inspection 5 (clean, used as unlabeled). We can see from Fig. \ref{fig:aco_count_unlabeled} that the validated $\mathscr{D}$, generally, predicts a decreasing number of anomalies along training and drops to 0 at some point. We stopped training right before the number of anomalies dropped to 0 and kept that $\mathscr{D}$ model for testing. On the other hand, F1 scores obtained from a labeled validation set went up first and peaked at some epoch values, then dropped down (see Fig. \ref{fig:f1_labeled}). Therefore, for testing, we selected the models at the epochs of peak F1. Table \ref{tab:model_selection} compares the average testing results of the two model selection methods based on the data from inspections 1 and 4 for three runs. We can see that, unsupervised  validation (by using unlabeled data) leads to a lower precision, while it produces a higher recall compared to the supervised validation (using labeled data). Generally, given the F1 scores, unsupervised validation relying on normal data is as effective as a supervised validation method with labeled data, which demonstrates the feasibility of the unsupervised model selection method proposed in Section \ref{sec:Discriminator model selection}.

\begin{table}[]
\centering
\caption{Average performance of model selection with labeled and unlabeled validation data ($wl$ = 100)}
\label{tab:model_selection}
\begin{tabular}{@{}cccccccccc@{}}
\toprule
$r_t$ (feet)             & \multicolumn{3}{c}{100}   & \multicolumn{3}{c}{150}   & \multicolumn{3}{c}{200}   \\ \midrule
Validation data & Recall & Precision & F1   & Recall & Precision & F1   & Recall & Precision & F1   \\ \midrule
Labeled             & 0.72   & 0.77      & 0.74 & 0.79   & 0.81      & 0.80 & 0.83   & 0.84      & 0.83 \\
Unlabeled           & 0.80   & 0.69      & 0.74 & 0.85   & 0.73      & 0.78 & 0.87   & 0.74      & 0.80 \\ \bottomrule
\end{tabular}
\end{table}

\subsubsection{Discussion}
The proposed framework with convolutional neural network architecture was shown to have a good performance for time series anomaly detection, reaching an F1 score of 0.83 in the railroad case study, by using data from only one channel of vertical acceleration time series. The framework, in its current form, could be adopted for multivariate times series datasets by processing different channels of the data into a concatenated sub-sequence of observations across multiple channels. Alternatively, in future investigations, a two-dimensional convolutional neural network architecture could be used to process stacked subsequences from different channels. 

Furthermore, results showed that factors such as GAN hyperparameters, architectures, sliding window sizes and model selection methods have considerable influence on the performance and the stability of the trained model considering the randomized initialization of parameters in the GAN model. The framework was demonstrated to be effective if the discriminator model is efficiently selected. As both supervised (labeled) and unsupervised (unlabeled) validation processes showed, finite and optimized training epochs and learning rates are needed to identify the better $\mathscr{D}$ model as an effective anomaly detector. Moreover, as noted, the clustering step appeared to be more effective for larger window sizes, for which more diverse patterns are observed and more balanced clusters could be formed. When clusters are unbalanced, the clustering step appears to be less effective. Therefore, although potential benefits and improvement in performance could be achieved in some problems, it should be noted that the use of this step adds to the computational cost of the framework considering the need for training multiple GAN models, as well as the computational steps for clustering. Further investigations into influencing factors that affect the performance, repeatability, and reliability of the trained anomaly detectors, as well as multi-threading strategies that improve the computational cost of training for large datasets, are among the future directions of this study.   

\section{Conclusion}
Among the machine learning approaches to tackle anomaly detection problems on time series data, generative adversarial network is attracting increasing attention in recent years. However, few efforts have been made to exploit the potential of an independent $\mathscr{D}$ to be a standalone anomaly detector. In this study, we proposed an unsupervised framework, DEGAN, which contains three main components: GAN training, $\mathscr{D}$ model selection, and anomaly detection. It makes use of the information contained in normal time series (processed using a sliding window method) during GAN training and $\mathscr{D}$ model selection, enabling a validated $\mathscr{D}$ model to be an independent anomaly detector after training for a given system. In the anomaly detection component, kernel density estimation is used to assign a probability distribution to the suspicious data points identified by the selected $\mathscr{D}$ model and identify the areas with the highest probability of being anomalies. We evaluated DEGAN using a real-world case study dataset that includes repeated inspection data from a Class I railroad. The results showed a promising performance, with a recall of 80\% and precision of 86\%, averaged over three evaluations. Moreover, several of the factors that affect the performance of the framework were investigated, including sliding window sizes, architectures and hyperparameters of GAN, clustering of time series,  $\mathscr{D}$ model configuration with labeled/ unlabeled data. The findings in this study show that (1) standalone validated discriminator models (trained to adapt to their specific context) have the potential to be effective anomaly detectors for time series data; (2) unsupervised training and validation of the framework using normal (clean) time series is as effective as using a supervised validation method; (3) the unsupervised validation seems to increase the recall while supervised validation improves precision of the anomaly detection process; (4) sliding window sizes, training epochs, GAN architectures and learning rates are important factors that may have a considerable impact on the performance of the framework.

\section*{Acknowledgments}
This material is based upon work supported by the Federal Railroad Administration’s Research, Development and Technology. Any opinions, findings, and conclusions or recommendations
expressed in this material are those of the authors and do not necessarily reflect the views of the Federal Railroad Administration. The authors appreciate the support from Virginia Tech's Railway Technologies Laboratory (RTL).

\bibliographystyle{unsrt}  
\bibliography{MyLibrary}

\end{document}